%% file: main.tex
\documentclass{article}
\pdfoutput=1

\usepackage[numbers]{natbib}

\usepackage[final]{neurips_2023}

\usepackage[utf8]{inputenc} 
\usepackage[T1]{fontenc}    
\usepackage{url}            
\usepackage{booktabs}       
\usepackage{amsfonts}       
\usepackage{nicefrac}       
\usepackage{microtype}      
\usepackage{xcolor}         

\usepackage{times}
\usepackage{epsfig}
\usepackage{amsmath}

\usepackage{array}
\usepackage{subcaption}
\usepackage{multirow,multicol}
\usepackage[flushleft]{threeparttable}
\usepackage{comment}
\usepackage{algorithmic}
\usepackage{booktabs}
\usepackage{bbm, dsfont}
\usepackage[font=small,labelfont=bf]{caption}

\usepackage{amssymb}
\usepackage{pifont}

\newcommand{\tablestyle}[2]{\setlength{\tabcolsep}{#1}\renewcommand{\arraystretch}{#2}\centering\footnotesize}

\newcommand{\app}{\raise.17ex\hbox{$\scriptstyle\sim$}}

\newlength\savewidth

\makeatletter\renewcommand\paragraph{\@startsection{paragraph}{4}{\z@}
  {.5em \@plus1ex \@minus.2ex}{-.5em}{\normalfont\normalsize\bfseries}}\makeatother

\def\tablecite#1#{%
  \def\pretablecite{#1}%
  \tableciteaux}
\def\tableciteaux#1{%
  \textsuperscript{\expandafter\originalcite\pretablecite{#1}}%
}

\usepackage{enumitem}
\usepackage{wrapfig}
\usepackage{lipsum}

\newcolumntype{H}{>{\setbox0=\hbox\bgroup}c<{\egroup}@{}}
\newcolumntype{a}{>{\columncolor{Gray}}c}

\usepackage{tabu}
\usepackage{nicematrix}

\definecolor{ForestGreen}{rgb}{0.13, 0.55, 0.13}
\definecolor{Green}{rgb}{0.0, 0.5, 0.0}
\definecolor{green(munsell)}{rgb}{0.0, 0.66, 0.47}
\definecolor{green(ryb)}{rgb}{0.4, 0.69, 0.2}
\definecolor{green(pigment)}{rgb}{0.0, 0.65, 0.31}
\definecolor{citecolor}{HTML}{0071bc}
\definecolor{GrayXMark}{gray}{0.7}

\usepackage{tabularx}
\usepackage[export]{adjustbox}

\definecolor{ForestGreen}{rgb}{0.13, 0.55, 0.13}
\definecolor{Green}{rgb}{0.0, 0.5, 0.0}
\definecolor{green(munsell)}{rgb}{0.0, 0.66, 0.47}
\definecolor{green(ryb)}{rgb}{0.4, 0.69, 0.2}
\definecolor{green(pigment)}{rgb}{0.0, 0.65, 0.31}

\newcommand{\plus}[1]{\small\bf\textcolor{Green}{#1}}

\newcolumntype{x}[1]{>{\centering\let\newline\\\arraybackslash\hspace{0pt}}p{#1}}
\definecolor{Gray}{gray}{0.9}

\usepackage{makecell}

\definecolor{ForestGreen}{rgb}{0.13, 0.55, 0.13}
\definecolor{Green}{rgb}{0.0, 0.5, 0.0}
\definecolor{green(munsell)}{rgb}{0.0, 0.66, 0.47}
\definecolor{green(ryb)}{rgb}{0.4, 0.69, 0.2}
\definecolor{green(pigment)}{rgb}{0.0, 0.65, 0.31}
\definecolor{citecolor}{HTML}{0071bc}
\definecolor{GrayXMark}{gray}{0.7}

\usepackage{tabularx}
\usepackage[export]{adjustbox}

\usepackage{hyperref}
\hypersetup{
    colorlinks,
    linkcolor={black},
    citecolor={black},
    urlcolor={black}
}
\usepackage[capitalize]{cleveref}
\crefname{section}{Sec.}{Secs.}
\Crefname{section}{Section}{Sections}
\Crefname{table}{Table}{Tables}
\crefname{table}{Table}{Tabs.}

\definecolor{ForestGreen}{rgb}{0.13, 0.55, 0.13}
\definecolor{Green}{rgb}{0.0, 0.5, 0.0}
\definecolor{green(munsell)}{rgb}{0.0, 0.66, 0.47}
\definecolor{green(ryb)}{rgb}{0.4, 0.69, 0.2}
\definecolor{green(pigment)}{rgb}{0.0, 0.65, 0.31}

\usepackage{colortbl}
\newcommand{\cmark}{\text{\ding{51}}}%
\newcommand{\cmarkColor}{\cellcolor{green!15}\text{\ding{51}}}%
\newcommand{\xmarkColor}{\cellcolor{red!15}\text{\ding{55}}}%
\newcommand{\astrmarkColor}{\cellcolor{gray!15}\text{*}}%
\newcommand{\daggerrmarkColor}{\cellcolor{gray!15}\text{\dag}}%
\usepackage{xspace}
\newcommand{\ours}{HIPIE\xspace}

\newcommand{\eg}{\textit{e.g.}\xspace}
\newcommand{\ie}{\textit{i.e.}\xspace}

\usepackage{color, colortbl}

\usepackage{mwe}
\usepackage{qtree}
\usepackage{wrapfig}
\usepackage{graphicx}

\title{Hierarchical Open-vocabulary Universal Image Segmentation}

\author{
\begin{tabular}{c}
Xudong Wang$^{1*}$ \quad  Shufan Li$^{1*}$ \quad  Konstantinos Kallidromitis$^{2*}$ \\
Yusuke Kato$^{2}$ \quad Kazuki Kozuka$^{2}$ \quad  Trevor Darrell$^{1}$
\end{tabular} \\
\quad $^{1}$Berkeley AI Research, UC Berkeley
\quad \quad $^{2}$Panasonic AI Research \\
\quad \small{project page: \hyperlink{http://people.eecs.berkeley.edu/~xdwang/projects/HIPIE}{http://people.eecs.berkeley.edu/$\sim$xdwang/projects/HIPIE}}
}

\begin{document}

\maketitle

\begin{abstract}
  Open-vocabulary image segmentation aims to partition an image into semantic regions according to arbitrary text descriptions.  
  However,  complex visual scenes can be naturally decomposed into simpler parts and abstracted at multiple levels of granularity,  introducing inherent segmentation ambiguity.
  Unlike existing methods that typically sidestep this ambiguity and treat it as an external factor, our approach actively incorporates a hierarchical representation encompassing different semantic-levels into the learning process. 
  We also propose a decoupled text-image fusion mechanism and representation learning modules for both ``things'' and ``stuff''.\footnote[1]{The terms \textit{things} (countable objects, typically foreground) and \textit{stuff} (non-object, non-countable, typically background)~\cite{adelson2001seeing} are commonly used to distinguish between objects that have a well-defined geometry and are countable, \eg people, cars, and animals, and surfaces or regions that lack a fixed geometry and are primarily identified by their texture and/or material, \eg the sky, road, and water body. \\
  *: equal contribution}
  Additionally, we systematically examine the differences that exist in the textual and visual features between these types of categories.
  Our resulting model, named \textbf{\ours}, tackles \textbf{HI}erarchical, o\textbf{P}en-vocabulary, and un\textbf{I}v\textbf{E}rsal segmentation tasks within a unified framework.
  Benchmarked on over 40 datasets, \eg, ADE20K, COCO, Pascal-VOC Part, RefCOCO/RefCOCOg, ODinW and SeginW, \ours achieves the state-of-the-art results at various levels of image comprehension, including semantic-level (\eg, semantic segmentation), instance-level (\eg, panoptic/referring segmentation and object detection), as well as part-level (\eg, part/subpart segmentation) tasks.
\end{abstract}

\input{sections/1_introduction}
\input{sections/2_related_works}
\input{sections/3_method}
\input{sections/4_results}
\input{sections/5_conclusion}


\input{sections/6_appendix}

\newpage
\clearpage

{\small
\bibliographystyle{abbrv}
\bibliography{references}
}

\end{document}

%% file: sections/1_introduction.tex
\def\figTeaser#1{
    \captionsetup[sub]{font=small}
    \begin{figure*}[#1]
      \centering
      \includegraphics[width=0.99\textwidth,scale=1]{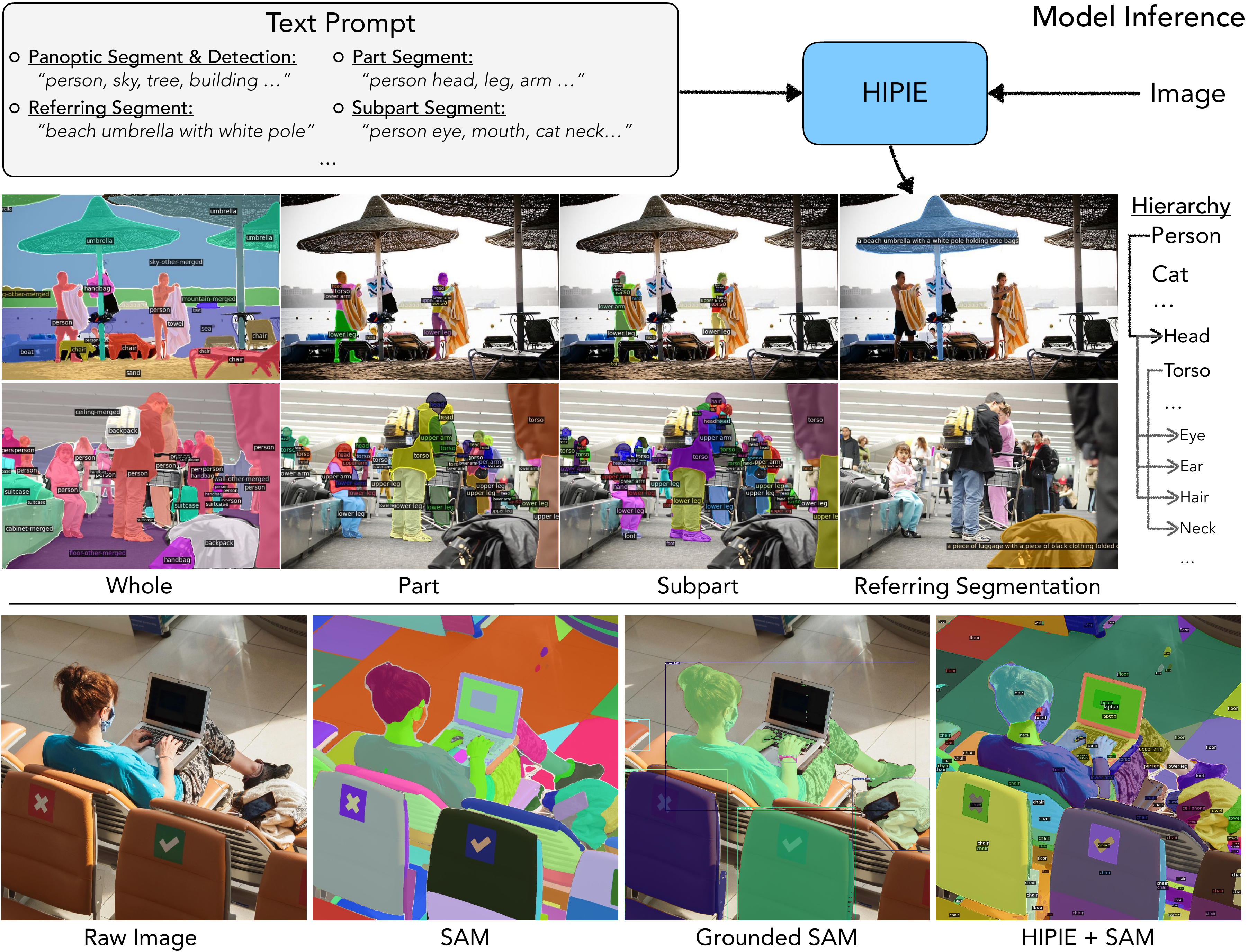}\vspace{-3pt}
      \caption{
      \ours is a  unified framework which, given an image and a set of arbitrary text descriptions, provides  hierarchical semantic, instance, part, and subpart-level image segmentations. 
      This includes open-vocabulary semantic (\eg, crowds and sky), instance/panoptic (\eg, person and cat), part (\eg, head and torso), subpart (\eg, ear and nose) and referring expression (\eg, umbrella with a white pole) masks. 
      \ours outperforms previous methods and established new SOTAs on these tasks regardless of their granularity or task specificity. 
      Bottom images: our method can seamlessly integrate with SAM to enable class-aware image segmentation on SA-1B.
      }
      \label{fig:teaser}
    \end{figure*}
}

\def\figAnalysis#1{
    \captionsetup[sub]{font=small}
    \begin{wrapfigure}{r}{0.5\textwidth}
      \begin{center}
      \vspace{-1.4\intextsep}
      \includegraphics[width=0.465\textwidth]{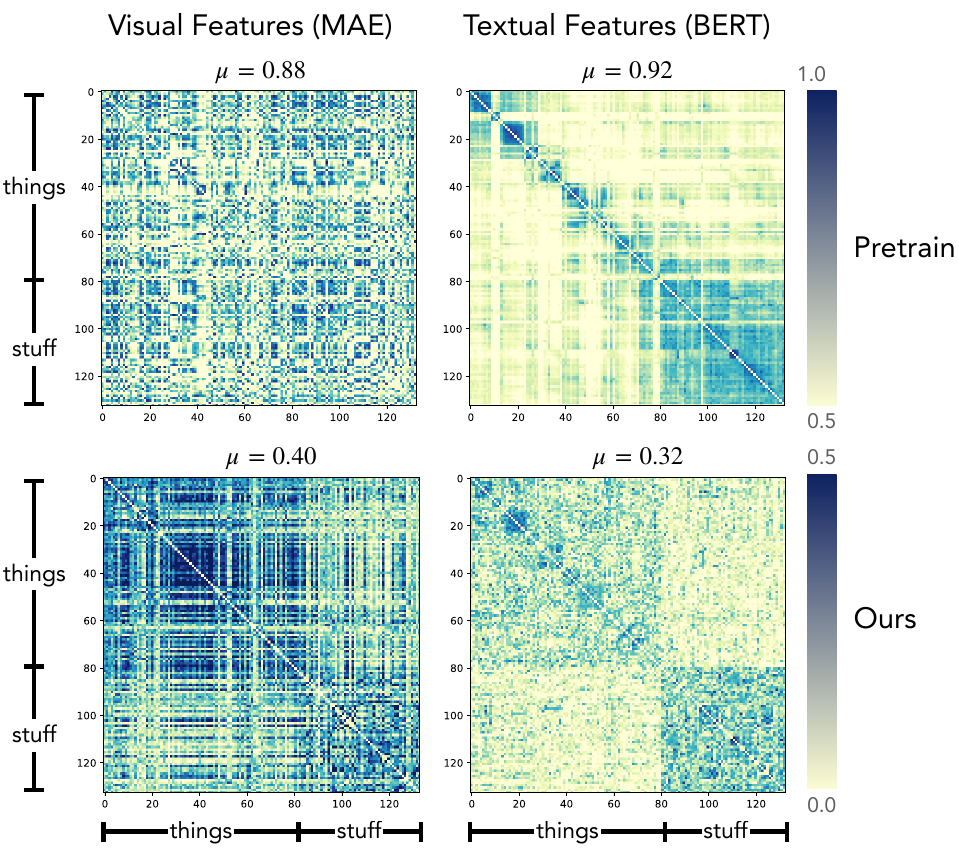}
      \end{center}
      \vspace{-10pt}
      \caption{
      Noticeable discrepancies exist in the between-class similarities of visual and textual features between stuff and thing classes.
      We propose a decoupled representation learning approach that effectively generates more discriminative visual and textual features.
      We extract similarity matrices for the visual features, obtained through a pretrained MAE~\cite{he2022masked} or our fine-tuned one, and for the text features, produced using a pretrained BERT~\cite{devlin2018bert} or fine-tuned one. 
      We report results on COCO-Panoptic~\cite{kirillov2019panoptic} and measure the mean similarity ($\mu$).
      }
      \vspace{-2.5\intextsep}
      \label{fig:analysis}
    \end{wrapfigure}
}

\section{Introduction} 
\label{sec:intro}

\figTeaser{h}

Image segmentation is a fundamental task in computer vision, enabling a wide range of applications such as object recognition, scene understanding, and image manipulation~\cite{szeliski2022computer,forsyth2002computer,nock2004statistical,dhanachandra2015image,long2015fully}.
Recent advancements in large language models pave the way for open-vocabulary image segmentation, where models can handle a wide variety of object classes using text prompts. 
However, there is no single ``correct'' way to segment an image. The inherent ambiguity in segmentation stems from the fact that the interpretations of boundaries and regions within an image depend on the specific tasks. 

Existing methods for open-vocabulary image segmentation typically address the ambiguity in image segmentation by considering it as an external factor beyond the modeling process. 
In contrast, we adopt a different approach by embracing this ambiguity and present \textbf{\ours}, as illustrated in \cref{fig:teaser}, a novel \textbf{HI}erarchical, o\textbf{P}en-vocabulary and un\textbf{I}v\textbf{E}rsal image segmentation and detection model.
This includes semantic-level segmentation, which focuses on segmenting objects based on their semantic meaning, as well as instance-level segmentation, which involves segmenting individual instances of objects or groups of objects (\eg, instance and referring segmentation). 
\figAnalysis{h}
Additionally, our model captures finer details by incorporating part-level segmentation, which involves segmenting object parts/subparts.
By encompassing  different granularity, HIPIE allows for a more comprehensive and nuanced analysis of images, enabling a richer understanding of their contents. 


To design \ours, we begin by  investigating the design choices for open-vocabulary image segmentation (OIS). Existing methods on OIS typically adopt a text-image fusion mechanism, and employ a shared representation learning module for both stuff and thing classes~\cite{cheng2021mask2former,zhang2022dino, yan2023universal,ding2022open,xu2023open}.  
\cref{fig:analysis} shows the similarity matrics of visual and textual features between stuff and thing classes. 
On this basis, we can derive several conclusions: 
\begin{itemize}[leftmargin=*,nosep]
    \item 
    Noticeable discrepancies exist in the between-class similarities of textual and visual features between stuff and thing classes. 
    \item 
    Stuff classes exhibit significantly higher levels of similarity in text features than things. 
\end{itemize}
This observation suggests that integrating textual features may yield more significant benefits in generating discriminative features for thing classes compared to stuff classes. Consequently, for thing classes, we adopt an early image-text fusion approach to fully leverage the benefits of discriminative textual features. Conversely, for stuff classes, we utilize a late image-text fusion strategy to mitigate the potential negative effects introduced by non-discriminative textual features.
Furthermore, the presence of discrepancies in the visual and textual features between stuff and thing classes, along with the inherent differences in their characteristics (stuff classes requiring better capture of texture and materials, while thing classes often having well-defined geometry and requiring better capture of shape information), indicates the need for decoupling the representation learning modules for producing masks for stuffs and things. 

In addition to instance/semantic-level segmentation, our model is capable of open-vocabulary hierarchical segmentation. 
Instead of treating part classes, like `dog leg', as standard multi-word labels, we concatenate class names from different granularity as prompts. During training, we supervise the classification head using both part labels, such as `tail', and instance labels, such as `dog', and we explicitly contrast a mask embedding with both instance-level and part-level labels.
In the inference stage, we perform two separate forward passes using the same image but different prompts to generate instance and part segmentation. 
This design choice empowers \textit{open-vocabulary} hierarchical segmentation, allowing us to perform part segmentation on novel part classes by randomly combining classes from various granularity, such as `giraffe' and `leg', even if they have never been seen during training.
By eliminating the constraints of predefined object classes and granularity, \ours offers a more flexible and adaptable solution for image segmentation.

We extensively benchmark \ours on various popular datasets to validate its effectiveness, including MSCOCO, ADE20K, Pascal Panoptic Part, and RefCOCO/RefCOCOg. \ours achieves state-of-the-art performance across all these datasets that cover a variety of tasks and granularity. 

To the best of our knowledge, \ours is the first hierarchical, open-vocabulary and universal image segmentation and detection model (see Table \ref{tab:distMethods}). By decoupling  representation learning and text-image fusion mechanisms for things vs. stuff classes, \ours overcomes the limitations of existing approaches and achieves state-of-the-art performance on various benchmarks.

%% file: sections/2_related_works.tex
\def\tabDistMethods#1{
    \begin{table}[#1]
    \centering
    \tablestyle{1.0pt}{1.2}
    \small
    \begin{tabular}{lx{1.2cm}x{1.35cm}x{1.35cm}x{1.35cm}x{1.35cm}x{1.6cm}x{1.6cm}x{1.35cm}}
    & \begin{tabular}[c]{@{}c@{}}Open \\Vocab. \end{tabular} & \begin{tabular}[c]{@{}c@{}}Instance \\Segment.\end{tabular} & \begin{tabular}[c]{@{}c@{}}Semantic \\Segment.\end{tabular} & \begin{tabular}[c]{@{}c@{}}Panoptic \\Segment.\end{tabular} & \begin{tabular}[c]{@{}c@{}}Referring \\Segment.\end{tabular} & \begin{tabular}[c]{@{}c@{}}Cls-agnostic \\Part Seg.\end{tabular} & \begin{tabular}[c]{@{}c@{}}Cls-aware \\Part Seg.\end{tabular} & \begin{tabular}[c]{@{}c@{}}Object \\Detection\end{tabular} \\
    \Xhline{0.8pt}
    SAM~\cite{kirillov2023segment} & \xmarkColor & \cmarkColor & \xmarkColor & \xmarkColor & \xmarkColor & \cmarkColor & \xmarkColor & \astrmarkColor\\
    SEEM~\cite{zou2023segment} & \cmarkColor &  \underline{\cmarkColor} & \cmarkColor & \underline{\cmarkColor} & \cmarkColor & \xmarkColor & \xmarkColor & \astrmarkColor\\
    ODISE~\cite{xu2023open} & \cmarkColor & \cmarkColor & \underline{\cmarkColor} & \cmarkColor & \xmarkColor & \xmarkColor & \xmarkColor & \astrmarkColor \\
    UNINEXT~\cite{yan2023universal} & \daggerrmarkColor & \cmarkColor & \xmarkColor & \xmarkColor & \underline{\cmarkColor} & \xmarkColor & \xmarkColor & \underline{\cmarkColor} \\
    X-Decoder~\cite{zou2022generalized} & \cmarkColor & \cmarkColor & \cmarkColor & \underline{\cmarkColor} & \cmarkColor & \xmarkColor & \xmarkColor & \astrmarkColor \\
    G-DINO~\cite{liu2023grounding} & \cmarkColor & \cmarkColor & \xmarkColor & \xmarkColor & \cmarkColor & \xmarkColor & \xmarkColor & \cmarkColor \\
    PPS~\cite{de2021part} & \xmarkColor & \xmarkColor & \xmarkColor & \xmarkColor & \xmarkColor & \cmarkColor & \underline{\cmarkColor} & \xmarkColor \\
    \Xhline{0.5pt}
    \ours & \cmarkColor & \cmarkColor & \cmarkColor & \cmarkColor & \cmarkColor & \cmarkColor & \cmarkColor & \cmarkColor \\
    \textit{vs. prev. SOTA} & - & \plus{+5.1} & \plus{+2.0} & \plus{+1.3} & \plus{+0.5} & - & \plus{+5.2} & \plus{+3.2} \\
    \end{tabular}
    \vspace{6pt}
    \caption{
    Our \ours is capable of performing all the listed segmentation and detection tasks and achieves the state-of-the-art performance using a unified framework.
    We present performance comparisons with SOTA methods on a range of benchmark datasets: AP$^{\text{mask}}$ for instance segmentation on MSCOCO~\cite{lin2014microsoft}, AP$^{\text{box}}$ for object detection on MSCOCO, oIoU for referring segmentation on RefCOCO+~\cite{yu2016modeling}, mIoU for semantic segmentation on Pascal Context\cite{zhou2019semantic}, and mIoU$_{\text{PartS}}$ for part segmentation on Pascal-Panoptic-Parts~\cite{de2021part}.
    The second best performing method for each task is \underline{underlined}.
    $^*$: object detection can be conducted via generating bounding boxes using instance segmentation masks. `Seg.' denotes segmentation. 
    \dag : In principle, UNINEXT can take arbitrary texts as labels, however, the original work focused on close-set performance and did not explore open-vocabulary inference. 
    }
    \label{tab:distMethods}
    \end{table}
}

\section{Related Works}
\label{sec:rworks}
\tabDistMethods{t!}

\noindent \textbf{{Open-Vocabulary Semantic Segmentation}}~\cite{bucher2019zero,xian2019semantic,li2022languagedriven,ghiasi2022scaling,rao2022denseclip,liang2022open,xu2022groupvit,xu2023learning,harary2022unsupervised} aims to segment an image into semantic regions indicated by text descriptions that may not have been seen during training. 
ZS3Net~\cite{bucher2019zero} combines a deep visual segmentation model with an approach to generate visual representations from semantic word embeddings to learn pixel-wise classifiers for novel categories.
LSeg~\cite{li2022languagedriven} uses CLIP's text encoder~\cite{radford2021learning} to generate the corresponding semantic class's text embedding, which it then aligns with the pixel embeddings.
OpenSeg~\cite{ghiasi2022scaling} adopts a grouping strategy for pixels prior to learning visual-semantic alignments. By aligning each word in a caption to one or a few predicted masks, it can scale-up the dataset and vocabulary sizes.
%
%
GroupViT~\cite{xu2022groupvit} is trained on a large-scale image-text dataset using contrastive losses. With text supervision alone, the model learns to group semantic regions together.
OVSegmentor~\cite{xu2023learning} uses learnable group tokens to cluster image pixels, aligning them with the corresponding caption embeddings. 

\noindent \textbf{{Open-Vocabulary Panoptic Segmentation}} (OPS) unifies semantic and instance segmentation, and aims to perform these two tasks for arbitrary categories of text-based descriptions during inference time~\cite{ding2022open,xu2023open,zou2022generalized,zou2023segment,yan2023universal}. 
MaskCLIP~\cite{ding2022open} first predicts class-agnostic masks using a mask proposal network. Then, it refines the mask features through Relative Mask Attention interactions with the CLIP visual model and integrates the CLIP text embeddings for open-vocabulary classification.
ODISE~\cite{xu2023open} unifies Stable Diffusion~\cite{rombach2022high}, a pre-trained text-image diffusion model, with text-image discriminative models, \eg CLIP~\cite{radford2021learning}, to perform open-vocabulary panoptic segmentation.
X-Decoder~\cite{zou2022generalized} takes two types of queries as input: generic non-semantic queries that aim to decode segmentation masks for universal segmentation, 
and textual queries to make the decoder language-aware for various open-vocabulary vision tasks. 
UNINEXT~\cite{yan2023universal} unifies diverse instance perception tasks into an object discovery and retrieval paradigm, enabling flexible perception of open-vocabulary objects by changing the input prompts.

\noindent \textbf{Referring Segmentation} learns valid multimodal features between visual and linguistic modalities to segment the target object described by a given natural language expression~\cite{hu2016segmentation,yu2018mattnet,hui2020linguistic,jing2021locate,feng2021encoder,yang2022lavt,wu2022towards,liu2023polyformer,zhao2023unleashing}.
It can be divided into two main categories: 
1) \textbf{\textit{Decoder-fusion}} based method~\cite{ding2021vision,wang2022cris,zhao2023unleashing,liu2023polyformer} first extracts vision features and language features, respectively, and then fuses them with a multi-modal design. 
2) \textbf{\textit{Encoder-fusion}} based method~\cite{feng2021encoder,yang2022lavt,li2021referring} fuses the language features into the vision features early in the vision encoder.

\textbf{Parts Segmentation} learns to segment instances into more fine-grained masks. PPP \cite{de2021part} established a baseline of hierarchical understanding of images by combining a scene-level panoptic segmentation model and part-level segmentation model. JPPF \cite{jagadeesh2022multijppf} improved this baseline by introducing joint Panoptic-Part Fusion module that achieves comparable performance with significantly smaller models. 

\noindent \textbf{Promptable Segmentation.} The Segment Anything Model (SAM)~\cite{kirillov2023segment} is an approach for building a fully automatic promptable image segmentation model that can incorporate various types of human interventions, such as texts, masks, and points.
SEEM~\cite{zou2023segment} proposes a unified prompting scheme that encodes user intents into prompts in a joint visual-semantic space. This approach enables SEEM to generalize to unseen prompts for segmentation, achieving open-vocabulary and zero-shot capabilities.
Referring segmentation can also be considered as promptable segmentation with  text prompts.

\textbf{Comparison with Previous Work.} Table~\ref{tab:distMethods} compares our \ours method with previous work in terms of key properties. Notably, \ours is the only method that supports open-vocabulary universal image segmentation and detection, enabling the object detection, instance-, semantic-, panoptic-, hierarchical-(whole instance, part, subpart), and referring-segmentation tasks, all within a single unified framework.

%% file: sections/3_method.tex
\def\figDiagram#1{
    \captionsetup[sub]{font=small}
    \begin{figure*}[#1]
      \centering
      \includegraphics[width=0.99\textwidth]{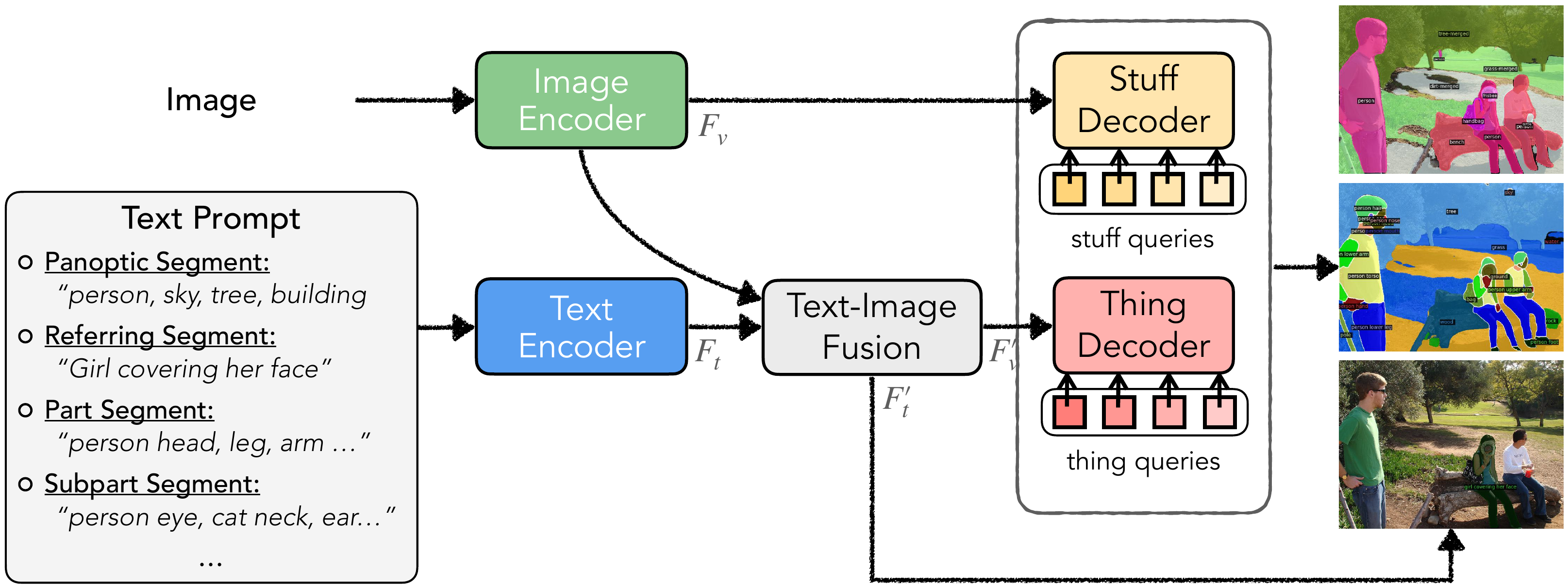}\vspace{-1pt}
      \caption{Diagram of \ours for hierarchical, universal and open-vocabulary image segmentation and detection. The image and text prompts are first passed to the image and text decoder to obtain visual features $F_v$  and text features $F_t$. Early fusion is then applied to merge image and text features to get $F_v', F_t'$. Two independent decoders are used for things (foreground) classes and stuff (background) classes. }
      \label{fig:diagram}
    \end{figure*}
}

\def\figDiagramDesignChoice#1{
    \captionsetup[sub]{font=small}
    \begin{figure*}[#1]
      \centering
      \includegraphics[width=0.99\textwidth]{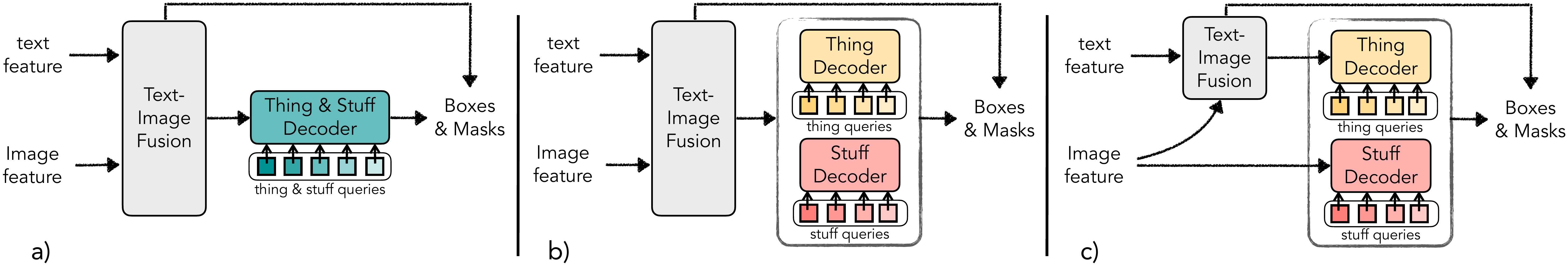}\vspace{-4pt}
      \caption{Various design choices for generating thing and stuff masks with arbitrary text descriptions. In version a), We use a single decoder for all masks. Early fusion is applied. In version b), two independent decoders are used for things and stuff classes. Early fusion is adopted for both decoders. Version c) is identical to version b) with the only difference being that the stuff decoder do not make use of early fusion. }
      \label{fig:diagram-design-choice}
    \end{figure*}
}

\def\figHierarchal#1{
    \captionsetup[sub]{font=small}
    \begin{figure*}[#1]
      \centering
      \includegraphics[width=0.95\textwidth]{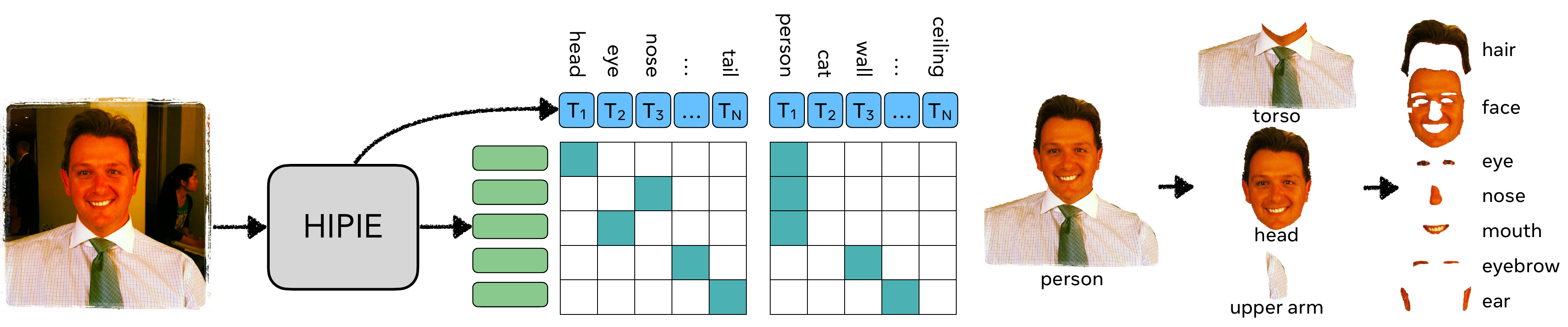}\vspace{-3pt}
      \caption{Hierarchical segmentation pipeline. We concatenate the instance class names and part class names as labels. During the training process, we supervise the classification head using both part labels and instance labels. During inference, we perform two separate forward passes using the same image but different prompts to generate instance and part segmentations. By combining the part segmentation and instance segmentation of the same image, we obtain hierarchical segmentation results on the right side.}
      \label{fig:hier}
    \end{figure*}
}

\figDiagram{t}

\section{Method}
We consider all relevant tasks under the unified framework of language-guided segmentation, which performs open-vocabulary segmentation and detection tasks for arbitrary text-based descriptions.

\subsection{Overall Framework}
\label{sec:framework}

The proposed \ours model comprises three main components, as illustrated in \cref{fig:diagram}:

\textit{1) Text-image feature extraction and information fusion (detailed in \cref{sec:text-prompts,sec:feature-extraction,sec:feature-fusion}):} We first generate a text prompt $T$ from labels or referring expressions. Then, we extract image ($I$) and text ($T$) features $F_v=\text{Enc}_v(I),F_t=\text{Enc}_t(T)$ using image encoder $\text{Enc}_v$ and text encoder $\text{Enc}_t$, respectively. We then perform feature fusion and obtain fused features $F_v', F_t'=\text{FeatureFusion}(F_v, F_t)$.

\textit{2) Foreground (referred to as things) and background (referred to as stuffs) mask generation (detailed in \cref{sec:mask-generation}):} Each of the decoders takes in a set of image features and text features and returns sets of masks, bounding boxes, and object embeddings $(M, B, E)$. We compute the foreground and background proposals and concatenate them to obtain the final proposals and masks as follows:
\begin{equation}
  \begin{array}{rlll}  
    \mathrm{Stuff:} & (M_2,B_2, E_2) & = & \mathrm{StuffDecoder}(F_v,F_t) \\
    \mathrm{Thing:} & (M_1,B_1, E_1) & = & \mathrm{ThingDecoder}(\mathrm{FeatureFusion}(F_v, F_t)) \\
    \mathrm{Overall:} & (M,B,E) & = & (M_1 \oplus M_2, B_1 \oplus B_2, E_1 \oplus E_2) \\
  \end{array}
\end{equation}
where $\oplus$ denotes the concatenation operation. 

\textit{3) Proposal and mask retrieval using text prompts (detailed in \cref{sec:open-vocabulary-seg}):}
To assign class labels to these proposals, we compute the cosine similarity between object embedding $E$ and the corresponding embedding $E_i'$ of class $i\in\{1,2...,c\}$. For a set of category names, the expression is a concatenated string containing all categories. We obtain $E_i'$ by pooling tokens corresponding to each label from the encoded sequence $F_t$. For referring expressions, we taken the [CLS] token from BERT output as $E_i'$.

\subsection{Text Prompts}
\label{sec:text-prompts}
Text prompting is a common approach used in open-vocabulary segmentation models~\cite{hu2016segmentation,yu2018mattnet,xu2022simple,yan2023universal}. 

For open-vocabualry instance segmentation, panoptic segmentation, and semantic segmentation, the set of all labels $C$ is concatenated into a single text prompt $T_i$ using a ``.'' delimiter. Given an image $I$ and a set of text prompts $T$, the model aims to classify $N$ masks in the label space $C\cup \{``other"\}$, where $N$ is the maximum number of mask proposals generated by the model.

For referring expressions, the text prompt is simply the sentence itself. The goal is to locate one mask in the image corresponding to the language expression.

\subsection{Image and Text Feature Extraction}
\label{sec:feature-extraction}

We employ a pretrained BERT model~\cite{devlin2018bert} to extract features for text prompts. Because the BERT-base model can only process input sequences up to 512 tokens, we divide longer sequences into segments of 512 tokens and encode each segment individually. The resulting features are then concatenated to obtain features of the original sequence length.

We utilize ResNet-50~\cite{he2016deep} and Vision Transformer (ViT)~\cite{dosovitskiyimage} as base architectures for image encoding. 
In the case of ResNet-50, we extract multiscale features from the last three blocks and denote them as $F_v$. For ViT, we use the output features from blocks 8, 16, and 32 as the multiscale features $F_v$.

\figDiagramDesignChoice{t}

\subsection{Text-Image Feature Fusion}
\label{sec:feature-fusion}
We explored several design choices for text-image feature fusion and mask generation modules as shown in \cref{fig:diagram-design-choice} and \cref{tab:ablation_arch}, and discovered that \cref{fig:diagram-design-choice}c) can give us the optimal performance. 
We adopt bi-directional cross-attention ($\mathrm{Bi\mbox{-}Xattn}$) to extract text-guided visual features $F_{t2v}$ and image-guided text features $F_{v2t}$. 
These attentive features are then integrated with the vanilla text features $F_t$ and image features $F_v$ through residual connections, as shown below:
\begin{equation}
  \begin{array}{lll}
    F_{t2v}, \: F_{v2t} & = & \mathrm{Bi\mbox{-}Xattn}(F_v,F_t) \\ 
    (F_v', \: F_t') & = & (F_v + F_{t2v}, \: F_t + F_{v2t})
  \end{array}
\end{equation}
where $F_v$ and $F_t$ represent the visual and text-prompt features, respectively. 

\subsection{Thing and Stuff Mask Generation}
\label{sec:mask-generation}
We then generate masks and proposals for the thing and stuff classes by utilizing $F_v'$ and $F_t'$ that we obtained in \cref{sec:feature-fusion}.

\textbf{Model Architecture.} While  architectures such as Mask2Former and MaskDINO~\cite{cheng2021mask2former,li2022mask} can perform instance, semantic and panoptic segmentation simultaneously, models trained jointly show inferior performance compared with the same model trained for a specific task (\eg instance segmentation only). 
We hypothesize that this may result from the different distribution of spatial location and geometry of foreground instance masks and background semantic masks. For example, instance masks are more likely to be connected, convex shapes constrained by a bounding box, whereas semantic masks may be disjoint, irregular shapes spanning across the whole image. 

To address this issue, in a stark contrast to previous approaches~\cite{yan2023universal,liu2023grounding,xu2022simple} that use a unified decoder all both stuffs and things, we decouple the stuff and thing mask prediction using two separate decoders. 
For the thing decoder, we adopt Deformable DETR~\cite{zhu2020deformable} with a mask head following the UNINEXT~\cite{yan2023universal} architecture and incorporate denoising procedures proposed by DINO~\cite{zhang2022dino}. For the stuff decoder, we use the architecture of MaskDINO~\cite{li2022mask}.

\textbf{Proposal and Ground-Truth Matching Mechanisms.} 
We make the following distinctions between the two heads. For thing decoder, we adopt simOTA~\cite{ge2021yolox} to perform many-to-one matching between box proposals and ground truth when calculating the loss. We also use box-iou-based NMS to remove duplicate predictions. 
For the stuff decoder, we adopt one-to-one Hungarian matching~\cite{kuhn1955hungarian}. Additionally, we disable the box loss for stuff masks. 
We set the number of queries to 900 for the things and 300 for the stuffs.

\textbf{Loss Functions.} 
For both decoders, we calculate the class logits as the normalized dot product between mask embeddings ($M$) and text embeddings ($F_t'$). 
We adopt Focal Loss~\cite{lin2017focal} for classification outputs, L1 loss, and GIoU loss~\cite{rezatofighi2019generalized} for box predictions, pixel-wise binary classification loss and DICE loss~\cite{sudre2017generalised} for mask predictions. 
Given predictions $ (M_1,B_1, E_1),(M_2,B_2, E_2)$, groundtruth labels $(M',B',C)$ and its foreground and background subset $(M_f',B_f',C_f)$ and $(M_b',B_b',C_b)$, The final Loss is computed as 
\begin{equation}
  \begin{array}{lll}
     \mathcal{L}_{\text{thing}}&=& \lambda_{\text{cls}} \mathcal{L}_{\text{cls}}(E_1,C_{f}') +\lambda_{\text{mask}}\mathcal{L}_{\text{mask}}(M_1,M_{f}') + \lambda_{\text{box}}\mathcal{L}_{\text{box}}(B_1,B_{f}') \\
     \mathcal{L}_{\text{stuff}}&=& \lambda_{\text{cls}} \mathcal{L}_{\text{cls}}(E_2,C') +\lambda_{\text{mask}}\mathcal{L}_{\text{mask}}(M_2,M') + \lambda_{\text{box}}\mathcal{L}_{\text{box}}(B_2,B_b') \\
     \mathcal{L} &=& \mathcal{L}_{\text{thing}}+ \mathcal{L}_{\text{stuff}} \\
    \end{array}
    \label{eq:loss}
\end{equation}
where $\mathcal{L}_{\text{box}} = \lambda_{L1} \mathcal{L}_{L1} + \lambda_{\text{giou}} \mathcal{L}_{\text{giou}}$, $\mathcal{L}_{\text{mask}} = \lambda_{\text{ce}}\mathcal{L}_{\text{ce}} + \lambda_{\text{dice}}\mathcal{L}_{\text{dice}}$, and $\mathcal{L}_{\text{cls}} = \mathcal{L}_{\text{focal}}$.
Note that while we do not use the stuff decoder for thing prediction, we still match its predictions with things and compute the class and box losses in the training. We find such auxiliary loss setup make the stuff decoder aware of the thing distribution and imporves the final performance. 

\subsection{Open-Vocabulary Universal Segmentation}
\label{sec:open-vocabulary-seg}
In closed set setting, we simply merge the output of two decoders and perform the standard postprocessing of UNINEXT~\cite{yan2023universal} and MaskDINO~\cite{li2022mask} to obtain the final output. 

In zero-shot open vocabulary setting, we follow ODISE~\cite{xu2023open} and combining our classification logits with a text-image discriminative model, \eg, CLIP~\cite{radford2021learning}. 
Specially, given the a mask $M$ on image $I$, its features $E$ and test classes $C_{\text{test}}$, we first compute the probability $p_1(E,C_{\text{test}})=\mathbb{P}(C_{\text{test}}|E)$ in the standard way as mentioned before. 
We additionally compute mask-pooled features of $M$ from the vision encoder $\mathcal{V}$ of CLIP as $E_{\text{CLIP}}=\text{MaskPooling}(M,\mathcal{V}(I))$. Then we compute the CLIP logits $p_2(E,C_{\text{test}})=\mathbb{P}(C_{\text{test}}|E_{\text{CLIP}})$  as the similarity between the CLIP text features and the $E_{\text{CLIP}}$. Finally we combine the final prediction as 
\begin{equation}
    p_{\text{final}}(E,C_{\text{test}})\propto p_1(E,C_{\text{\text{test}}})^\lambda p_2(E,C_{\text{test}})^{1-\lambda}
    \label{eq:open_vocab}
\end{equation}
Where $\lambda$ is a balancing factor. Emprically, we found such setting leads to better performance than naively relying completely on CLIP features only or close-set logits. 

\figHierarchal{}

\subsection{Hierarchical segmentation}
In addition to the instance-level segmentation, we can also perform part-aware hierarchical segmentation.
We concatenate the instance class names and part class names as labels. Some examples are "human ear", and "cat head". In the training process, we supervise the classification head with both part labels and instance labels. Specifically, we replace $L_{cls}$ with $L_{clsPart}+L_{clsThing}$ in \cref{eq:loss}.   We combine parts segmentation and instance segmentation of the same image to get part-aware instance segmentation. Additionally, layers of hierarchy is obtained by grouping the parts. For example, the "head" consists of ears, hair, eyes, nose, etc. \cref{fig:hier} illustrates this process.  \cref{fig:hierarchal} highlights the difference of our approach with other methods.

\subsection{Class-aware part segmentation with SAM}
\label{sec:hie-seg}
We can also perform the class-aware hierarchical segmentation by combining our semantic output with class-agnostic masks produced by SAM~\cite{kirillov2023segment}.
Specifically, given semantic masks $M$, their class probability $P_M$, and SAM-generated part masks $S$, we compute the class probability of mask $S_i \in S$ with respect to class $j$ as
\begin{equation}    
    P_S(S_i,j)\propto \sum_{M_k \in M} P_M( M_k,j) | M_k \cap  S_i |
\end{equation}
Where $ | M_k \cap  S_i |$ is the area of intersection between mask $M_k$ and $S_i$. We combine our semantic output with SAM because our pretraining datasets only contains object-centric masks, whereas the SA-1B dataset used by SAM contains many local segments and object parts.
\label{sec:method}

%% file: sections/4_results.tex
\def\tabPanopticSeg#1{
\begin{table}[#1]
    \tablestyle{1.9pt}{1.0}
    \begin{center}
    \begin{tabular}{p{2.2cm}lHcHcccclcccclc}
    \multirow{2}{*}{Method} & \multirow{2}{*}{Backbone} & \multirow{2}{*}{\#Params} & && \multicolumn{4}{c}{{COCO}} && \multicolumn{4}{c}{{ADE20K}} &&  \multicolumn{1}{c}{{PAS-P}} \\
    \cline{6-9} \cline{11-14} \cline{16-16}
    &&&&& \multicolumn{1}{c}{PQ} & \multicolumn{1}{c}{AP$^{\text{mask}}$} & \multicolumn{1}{c}{AP$^{\text{box}}$} & \multicolumn{1}{c}{mIoU} && \multicolumn{1}{c}{PQ} & \multicolumn{1}{c}{AP$^{\text{mask}}$} & \multicolumn{1}{c}{AP$^{\text{box}}$} & \multicolumn{1}{c}{mIoU} &&  \multicolumn{1}{c}{mIoU$_\text{PartS}$}\\ [.1em]
    \Xhline{0.8pt}
    MaskCLIP \cite{ding2022open} & ViT16 & N/A &&& \phantom{1}\phantom{1}- & \phantom{1}\phantom{1}- & \phantom{1}\phantom{1}- & \phantom{1}\phantom{1}- && 15.1 & \phantom{1}6.0 & \phantom{1}\phantom{1}- & 23.7  && - \\
    X-Decoder \cite{zou2022generalized} & FocalT & N/A &&& 52.6 & 41.3 & \phantom{1}\phantom{1}- & 62.4 && 18.8 & \phantom{1}9.8 & \phantom{1}\phantom{1}- & 25.0 && - \\
    X-Decoder & DaViT-B & N/A &&& 56.2 & 45.8 & \phantom{1}\phantom{1}- & 66.0 && 21.1 & 11.7 & \phantom{1}\phantom{1}- & 27.2 && - \\
    SEEM \cite{zou2023segment}& FocalT & N/A &&& 50.6 & 39.5 & \phantom{1}\phantom{1}- & 61.2 && \phantom{1}\phantom{1}- & \phantom{1}\phantom{1}- & \phantom{1}\phantom{1}- & \phantom{1}\phantom{1}- && - \\
    SEEM & DaViT-B & N/A &&& 56.2 & 46.8 & \phantom{1}\phantom{1}- & 65.3 && \phantom{1}\phantom{1}- & \phantom{1}\phantom{1}- & \phantom{1}\phantom{1}- & \phantom{1}\phantom{1}-  && - \\
    ODISE \cite{xu2023open} & ViT-H+SD & 1521M &&& 55.4 & 46.0 & 46.1 & 65.2 && \textbf{22.6 }& 14.4  & 15.8 & \textbf{29.9}  && - \\
    JPPF  \cite{jagadeesh2022multijppf}&  EffNet-b5  &  &&& - &- & \phantom{1}\phantom{1}- & - &&-& - & \phantom{1}\phantom{1}- & -&&  54.4 \\
    PPS  \cite{de2021part}&  RNST269  &  &&& - &- & \phantom{1}\phantom{1}- & - &&-& - & \phantom{1}\phantom{1}- & -&&  58.6 \\
    \rowcolor{Gray!60}
    \ours & RN50 & COCO && & 52.7 & 45.9 & 53.9 & 59.5 && 18.4 & 13.0 & 16.2 & 26.8  && 57.2 \\
    \rowcolor{Gray!60}
    \ours & ViT-H & TBD && &\textbf{58.0}&\textbf{51.9}&\textbf{61.3}&\textbf{66.8} &  & 20.6 & \textbf{15.0} &\textbf{ 18.7} & 29.0 && \textbf{63.8} \\
    \end{tabular}
    \end{center}
    \caption{Open-vocabulary panoptic segmentation (PQ), instance segmentation (AP$^{\text{mask}}$), semantic segmentation (mIoU), part segmentation (mIoU$_\text{PartS}$), and object detection (AP$^{\text{box}}$). N/A: not applicable. -: not reported.}
    \label{tab:panoptic-seg}
    \end{table}
}

\def\tabPanopticSegPoster#1{
\begin{table}[#1]
    \tablestyle{1.9pt}{1.0}
    \begin{center}
    \begin{tabular}{p{2.2cm}HHHHcccclcccclcccccccccc}
    \multirow{2}{*}{Method} & \multirow{2}{*}{Backbone} & \multirow{2}{*}{\#Params} & && \multicolumn{4}{c}{{COCO}} && \multicolumn{4}{c}{{ADE20K}} &&  \multicolumn{1}{c}{{PAS-P}}&& \multicolumn{3}{c}{RefCOCO/+/g} && A-847  \\
    \cline{6-9} \cline{11-14} \cline{16-16} \cline{18-20} \cline{22-22}
    &&&&& \multicolumn{1}{c}{PQ} & \multicolumn{1}{c}{AP$^{\text{mask}}$} & \multicolumn{1}{c}{AP$^{\text{box}}$} & \multicolumn{1}{c}{mIoU} && \multicolumn{1}{c}{PQ} & \multicolumn{1}{c}{AP$^{\text{mask}}$} & \multicolumn{1}{c}{AP$^{\text{box}}$} & \multicolumn{1}{c}{mIoU} &&  \multicolumn{1}{c}{mIoU$_\text{PartS}$} && \multicolumn{3}{c}{oIoU} && \multicolumn{1}{c}{mIoU} & \\ [.1em]
    \Xhline{0.8pt}
  
    \rowcolor{Gray!60}
    prev. SOTA &  & COCO && & 56.2 & 46.8 & 46.1& 66.0 && 22.6 & 14.4 & 15.8 & 29.9 && 58.6 &&  82.2 & 72.5 & 74.7 && 9.2\\
    \rowcolor{Gray!60}
    \ours & ViT-H & TBD && &\textbf{58.0}&\textbf{51.9}&\textbf{61.3}&\textbf{66.8} &  &\textbf{ 22.9} & \textbf{19.0} &\textbf{ 22.9} & 29.0 && \textbf{63.8} && \textbf{82.6 }&\textbf{ 73.0} & \textbf{75.3} && \textbf{9.7}\\
    \end{tabular}
    \end{center}
    \caption{Open-vocabulary panoptic segmentation (PQ), instance segmentation (AP$^{\text{mask}}$), semantic segmentation (mIoU), part segmentation (mIoU$_\text{PartS}$), and object detection (AP$^{\text{box}}$). N/A: not applicable. -: not reported.}
    \label{tab:panoptic-seg}
    \end{table}
}

\def\tabOpenSeg#1{
\begin{table}[#1]
    \tablestyle{1.9pt}{1.0}
    \begin{center}
    \begin{tabular}{p{2.2cm}Hcccclcccccc}
\multirow{2}{*}{Method} & \multirow{2}{*}{Venue} & \multirow{2}{*}{Data} &  \multicolumn{4}{c}{{A-150}} && A-847  && CTX459  && SeginW  \\
\cline{4-7} \cline{9-9} \cline{11-11} \cline{13-13}
 &&& PQ & AP$^{\text{mask}}$ & AP$^{\text{box}}$ & mIoU && mIoU && mIoU && AP$^{\text{mask}}$ \\

    \Xhline{0.8pt}
    
OpenSeed              & ICCV2023 & O365,COCO                                  & 19.7 & 15.0 & 17.7 & 23.4 && -   && -    && 36.1 \\
\multirow{1}{*}{X-Decoder}              & \multirow{1}{*}{CVPR2023 } & \tiny COCO,CC3M,SBU-C,VG,COCO-Caption,(Florence) & \multirow{1}{*}{21.8} & \multirow{1}{*}{13.1} & \multirow{1}{*}{-}    & \multirow{1}{*}{29.6} && \multirow{1}{*}{9.2} && \multirow{1}{*}{16.1} && \multirow{1}{*}{32.2} \\ 

UNINEXT               & CVPR2023 & O365,COCO,RefCOCO                          & 8.9  & 14.9 & 11.9 & 6.4  && 1.8 && 5.8  && 42.1 \\
HIPIE w/o CLIP  & -        & O365,COCO,RefCOCO,PACO                          & 18.1 & 16.7 & 20.2 & 19.8 && 4.8 && 12.2 && 41.0 \\
HIPIE w/ CLIP   & -        & + (CLIP)                  & 22.9 & 19.0 & 22.9 & 29.0 && 9.7 && 14.4 && 41.6 
\end{tabular}
    \end{center}
    \caption{Open-Vocabulary Universal Segmentation. We compare against other universal multi-task segmentation models. (*) denotes pretraining dataset of representations. }
    \label{tab:panoptic-seg}
    \end{table}
}

\def\tabDetInstRef#1{
\begin{table}[#1]
    \parbox{.5\textwidth}{
        \tablestyle{1.pt}{1.0}
        \begin{center}
        \begin{tabular}{p{2.8cm}lHHHcHcccHHHHHH}
        \multirow{2}{*}{Method} & \multirow{2}{*}{Backbone} & \multirow{2}{*}{\#Params} & \multirow{2}{*}{Data} && \multicolumn{5}{c}{{Object Detection}} \\
        \cline{6-10}
        &&&&& \multicolumn{1}{c}{AP} & & \multicolumn{1}{c}{AP$_{S}$} & \multicolumn{1}{c}{AP$_{M}$} & \multicolumn{1}{c}{AP$_{L}$} \\ [.1em]
        \Xhline{0.8pt}
        Deform. DETR~\cite{zhu2020deformable} & RN50 & N/A &&& 46.9 & 65.6 & 29.6 & 50.1 & 61.6 &&  \\
        DN-DETR~\cite{li2022dn} & RN50 & N/A &&& 48.6 & 67.4 & 31.0 & 52.0 & 63.7 && \\
        UNINEXT~\cite{yan2023universal} & RN50 & N/A &&& 51.3 & 68.4 & 32.6 & 55.7 & 66.5 && 44.9 & 67.0 & 26.3 & 48.5 & 59.0 \\
        \rowcolor{Gray!60}
        \ours & RN50 & COCO && & 53.9 &70.9&37.5&58.0&68.0 && 45.8 &68.8&26.9&50.0& 63.1\\
        \rowcolor{Gray!60}
        \textit{vs. prev. SOTA} &&&&& \plus{+2.6} & \plus{+2.5} & \plus{+4.9} & \plus{+2.3} & \plus{+1.5} && \plus{+0.9} & \plus{+1.8} & \plus{+0.6} & \plus{+1.5} & \plus{+4.1} \\
        \Xhline{0.8pt}
        Cas. Mask-RCNN~\cite{cai2019cascade} & CNeXtL & N/A &&& 54.8 & 73.8 & \phantom{1}\phantom{1}- & \phantom{1}\phantom{1}- & \phantom{1}\phantom{1}- && 47.6 & 71.3 & & & \\
        ViTDet-H~\cite{li2022exploring} & ViT-H & N/A &&& 58.7 & \phantom{1}\phantom{1}- & \phantom{1}\phantom{1}- & \phantom{1}\phantom{1}- & \phantom{1}\phantom{1}- && 50.9 & & & & \\
        UNINEXT~\cite{yan2023universal} & ViT-H & N/A &&& 58.1 & 74.9 & 40.7 & 62.5 & 73.6 && 51.8 & 76.2 & 33.3 & 55.9 & 67.5 \\
        \rowcolor{Gray!60}
        \ours & ViT-H & N/A &&& 61.3&76.5 &45.8 & 65.7& 75.9 && 51.9&76.9&33.6 & 56.0&70.3 \\
        \rowcolor{Gray!60}
        \textit{vs. prev. SOTA} &&&&& \plus{+3.2} & \plus{+1.6} & \plus{+5.1} & \plus{+3.2} & \plus{+2.3} && \plus{} & \plus{} & \plus{} & \plus{} & \plus{} \\
        \end{tabular}
        \end{center}
        \label{tab:det-inst-seg}
        \caption{Comparisons on the instance segmentation and object detection tasks. We evaluate model performance on the validation set of MSCOCO.}
    }
    \hfill
    \parbox{.48\textwidth}{
        \tablestyle{1.pt}{1.0}
        \begin{center}
        \begin{tabular}{p{1.9cm}lHHlHcHlHcHlHcH}
        \multirow{2}{*}{Method} & \multirow{2}{*}{Backbone} & \multirow{2}{*}{\#Params} & \multirow{2}{*}{Data} && \multicolumn{3}{c}{{COCO}} && \multicolumn{3}{c}{{COCO+}} && \multicolumn{3}{c}{{COCOg}} \\
        \cline{6-8} \cline{10-12} \cline{14-16}
        &&&&& \multicolumn{1}{c}{} & \multicolumn{1}{c}{oIoU} & && \multicolumn{1}{c}{} & \multicolumn{1}{c}{oIoU} & && \multicolumn{1}{c}{} & \multicolumn{1}{c}{oIoU} & \\ [.1em]
        \Xhline{0.8pt}
        MAttNet~\cite{yu2018mattnet} & RN101 & - &&& \phantom{1}\phantom{1}- &56.5 & 76.7 && \phantom{1}\phantom{1}- & 46.7 & 65.3 && \phantom{1}\phantom{1}- & 47.6 & 66.6 \\
        VLT~\cite{ding2022vlt} & Dark56 & - &&& \phantom{1}\phantom{1}- & 65.7 & 76.2 && \phantom{1}\phantom{1}- & 55.5 & 64.2 && \phantom{1}\phantom{1}- & 53.0 & 61.0 \\
        RefTR~\cite{muchen2021referring} & RN101 & - &&& \phantom{1}\phantom{1}- & 74.3 & 85.7 && \phantom{1}\phantom{1}- & 66.8 & 77.6 && \phantom{1}\phantom{1}- & 64.7 & 82.7 \\
        UNINEXT~\cite{yan2023universal} & RN50 & - &&& \phantom{1}\phantom{1}- & 77.9 & 89.7 && \phantom{1}\phantom{1}- & 66.2 & 79.7 && \phantom{1}\phantom{1}- & 70.0 & 84.0 \\
        UNINEXT~\cite{yan2023universal} & ViT-H & - &&& \phantom{1}\phantom{1}- & 82.2 & 92.6 && \phantom{1}\phantom{1}- & 72.5 & 85.2 && \phantom{1}\phantom{1}- & 74.7 & 88.7 \\
        \rowcolor{Gray!60}
        \ours & RN50 & TBD &&   && 78.3 &   90.1 & && 66.2 & 80.0 & & & 69.8  &83.6  \\
        \rowcolor{Gray!60}
        \ours & ViT-H & TBD && & &82.6& 93.0&& &73.0&85.5 & &&75.3&88.9 \\
        \rowcolor{Gray!60}
        \multicolumn{2}{l}{\textit{vs. prev. SOTA}} &&& & \plus{+0.4} & \plus{+0.4} & & & & \plus{+0.5} & & & \plus{} & \plus{+0.6} & \plus{} \\
        \end{tabular}
        \end{center}
        \caption{Comparison on the referring image segmentation (RIS) task. We evaluate the model performance on the validation sets of RefCOCO, RefCOCO+, and RefCOCOg datasets using overall IoU (oIoU) metrics.}
        \label{tab:refer-seg}
    }
\end{table}
}

\def\tabDetInst#1{
    \begin{table}[#1]
        \tablestyle{1.9pt}{1.0}
        \begin{center}
        \begin{tabular}{p{3.2cm}lHHlcccccHHHHHH}
        \multirow{2}{*}{Method} & \multirow{2}{*}{Backbone} & \multirow{2}{*}{\#Params} & \multirow{2}{*}{Data} && \multicolumn{5}{c}{{Object Detection}} \\
        \cline{6-10}
        &&&&& \multicolumn{1}{c}{AP} & \multicolumn{1}{c}{AP$_{50}$} & \multicolumn{1}{c}{AP$_{S}$} & \multicolumn{1}{c}{AP$_{M}$} & \multicolumn{1}{c}{AP$_{L}$} \\ [.1em]
        \Xhline{0.8pt}
        Faster R-CNN & RN50 & 1521M &&& 42.0 & 62.1 & 26.6 & 45.4 & 53.4 && \\
        Cascade Mask R-CNN & RN50 & N/A &&& 46.3 & 64.3 &&& && 38.6 & 60.0 & 21.7 & 40.8 & 49.6\\
        Deformable-DETR & RN50 & N/A &&& 46.9 & 65.6 & 29.6 & 50.1 & 61.6 &&  \\
        DN Deformable-DETR & RN50 & N/A &&& 48.6 & 67.4 & 31.0 & 52.0 & 63.7 && \\
        QueryInst & RN50 & N/A &&& & & & & && 40.6 & 63.0 & 23.4 & 42.5 & 52.8 \\
        UNINEXT~\cite{yan2023universal} & RN50 & N/A &&& 51.3 & 68.4 & 32.6 & 55.7 & 66.5 && 44.9 & 67.0 & 26.3 & 48.5 & 59.0 \\
        \rowcolor{Gray!60}
        \ours & RN50 & COCO && & 53.9 &70.9&37.5&58.0&68.0 && 45.8 &68.8&26.9&50.0& 63.1\\
        \rowcolor{Gray!60}
        \textit{vs. prev. SOTA} &&&&& \plus{+2.6} & \plus{+2.5} & \plus{+4.9} & \plus{+2.3} & \plus{+1.5} && \plus{+0.9} & \plus{+1.8} & \plus{+0.6} & \plus{+1.5} & \plus{+4.1} \\
        \Xhline{0.8pt}
        Cascade Mask R-CNN & ConvNeXt-L & N/A &&& 54.8 & 73.8 & & & && 47.6 & 71.3 & & & \\
        ViTDet-H & ViT-H & N/A &&& 58.7 & & & & && 50.9 & & & & \\
        Mask2Former & Swin-L & N/A &&& & & & & && 50.1 & & 29.9 & 53.9 & 72.1 \\
        UNINEXT~\cite{yan2023universal} & ViT-H & N/A &&& 58.1 & 74.9 & 40.7 & 62.5 & 73.6 && 51.8 & 76.2 & 33.3 & 55.9 & 67.5 \\
        \rowcolor{Gray!60}
        \ours & ViT-H & N/A &&& 61.3&76.5 &45.8 & 65.7& 75.9 && 51.9&76.9&33.6 & 56.0&70.3 \\
        \rowcolor{Gray!60}
        \textit{vs. prev. SOTA} &&&&& \plus{+3.2} & \plus{+1.6} & \plus{+5.1} & \plus{+3.2} & \plus{+2.3} && \plus{} & \plus{} & \plus{} & \plus{} & \plus{} \\
        \end{tabular}
        \end{center}
        \caption{Comparions with state-of-the-art methods on the close-set instance segmentation and object detection. We evaluate model performance on the validation set of MSCOCO. }
        \label{tab:det-inst-seg}
    \end{table}
}

\def\tabReferSeg#1{
    \begin{table}[#1]
    \tablestyle{1.9pt}{1.0}
    \begin{center}
    \begin{tabular}{p{2.2cm}lHHlHcclHcclHcc}
    \multirow{2}{*}{Method} & \multirow{2}{*}{Backbone} & \multirow{2}{*}{\#Params} & \multirow{2}{*}{Data} && \multicolumn{3}{c}{{RefCOCO}} && \multicolumn{3}{c}{{RefCOCO+}} && \multicolumn{3}{c}{{RefCOCOg}} \\
    \cline{6-8} \cline{10-12} \cline{14-16}
    &&&&& \multicolumn{1}{c}{} & \multicolumn{1}{c}{oIoU} & \multicolumn{1}{c}{Prec@.5} && \multicolumn{1}{c}{} & \multicolumn{1}{c}{oIoU} & \multicolumn{1}{c}{Prec@.5} && \multicolumn{1}{c}{} & \multicolumn{1}{c}{oIoU} & \multicolumn{1}{c}{Prec@.5} \\ [.1em]
    \Xhline{0.8pt}
    MAttNet & RN101 & - &&& \phantom{1}\phantom{1}- &56.5 & 76.7 && \phantom{1}\phantom{1}- & 46.7 & 65.3 && \phantom{1}\phantom{1}- & 47.6 & 66.6 \\
    VLT & Darknet56 & - &&& \phantom{1}\phantom{1}- & 65.7 & 76.2 && \phantom{1}\phantom{1}- & 55.5 & 64.2 && \phantom{1}\phantom{1}- & 53.0 & 61.0 \\
    RefTR & RN101 & - &&& \phantom{1}\phantom{1}- & 74.3 & 85.7 && \phantom{1}\phantom{1}- & 66.8 & 77.6 && \phantom{1}\phantom{1}- & 64.7 & 82.7 \\
    UNINEXT & RN50 & - &&& \phantom{1}\phantom{1}- & 77.9 & 89.7 && \phantom{1}\phantom{1}- & 66.2 & 79.7 && \phantom{1}\phantom{1}- & 70.0 & 84.0 \\
    \rowcolor{Gray!60}
    \ours & RN50 & TBD &&   && 78.3 &   90.1 & && 66.1 & 80.0 & & & 69.8  &83.6  \\
    \rowcolor{Gray!60}
    \textit{vs. prev. SOTA} &&&& \plus{} & \plus{} & \plus{} & \plus{} & \plus{} & \plus{} & \plus{} & \plus{} & & \plus{} & \plus{} & \plus{} \\
    \Xhline{0.8pt}
    UNINEXT & ViT-H & - &&& \phantom{1}\phantom{1}- & 82.2 & 92.6 && \phantom{1}\phantom{1}- & 72.5 & 85.2 && \phantom{1}\phantom{1}- & 74.7 & 88.7 \\
    \rowcolor{Gray!60}
    \ours & ViT-H & TBD && & &82.6& 93.0&& &73.0&85.5 & &&75.3&88.9 \\
    \rowcolor{Gray!60}
    \textit{vs. prev. SOTA} &&&& \plus{} & \plus{} & \plus{} & \plus{} & \plus{} & \plus{} & \plus{} & \plus{} & & \plus{} & \plus{} & \plus{} \\
    \end{tabular}
    \end{center}
    \caption{Comparison on the referring expression comprehension (REC), and referring image segmentation (RIS) tasks. The evaluation is carried out on the validation sets of RefCOCO, RefCOCO+, and RefCOCOg datasets using Precision@0.5 and overall IoU (oIoU) metrics for REC and RIS, respectively.}
    \label{tab:refer-seg}
    \end{table}
}

\def\tabSemanticSeg#1{
    \begin{table}[#1]
    \vspace{-15pt}
    \tablestyle{1.9pt}{1.0}
    \parbox{.46\textwidth}{
    \begin{center}
    \begin{tabular}{p{2.2cm}HHHHcccc}
    \multirow{1}{*}{Method} & \multirow{1}{*}{Backbone} & \multirow{1}{*}{Data} & A-847 & PC-459 & A-150 & PC-59 & PAS-21 & COCO \\
    \Xhline{0.8pt}
    ZS3Net~\cite{bucher2019zero} & RN101 & VOC & \phantom{1}\phantom{1}- & \phantom{1}\phantom{1}- & \phantom{1}\phantom{1}- & 19.4 & 38.3 & \phantom{1}\phantom{1}- \\
    LSeg+~\cite{li2022languagedriven,ghiasi2022scaling} & RN101 & COCO & \phantom{1}3.8 & \phantom{1}7.8 & 18.0 & 46.5 & \phantom{1}\phantom{1}- & 55.1 \\
    \rowcolor{Gray!60}
    \ours & RN50 & COCO & \phantom{1}\phantom{1}- & 10.9 & 26.8 & 53.6 & 75.7 & 59.5 \\
    \rowcolor{Gray!60}
    \textit{vs. prev. SOTA} &&&& \plus{+3.1} & \plus{+7.1} & \plus{+10.7} & \plus{+28.3} & \plus{+4.4} \\
    \Xhline{0.8pt}
    GroupViT~\cite{xu2022groupvit} & ViT-B & GCC+YFCC & \phantom{1}4.3 & \phantom{1}4.9 & 10.6 & 25.9 & 50.7 & 21.1 \\
    OpenSeg~\cite{ghiasi2022scaling} & eff-b7 & COCO & \phantom{1}6.3  & \phantom{1}9.0 & 21.1 & 42.1 & \phantom{1}\phantom{1}- & 36.1 \\
    MaskCLIP~\cite{ding2022open} & ViT-B & COCO & 
    \phantom{1}8.2 & 10.0 & 23.7 & 45.9 & \phantom{1}\phantom{1}- & \phantom{1}\phantom{1}- \\
    ODISE~\cite{xu2023open} & ViT-H+SD & COCO & 11.1 & 14.5 & 29.9 & 57.3 & 84.6 & 65.2 \\
    \rowcolor{Gray!60}
    \ours & ViT-H & COCO && 12.4 &29.0&59.3&83.3&66.8 \\
    \rowcolor{Gray!60}
    \textit{vs. prev. SOTA}  &&&& & -0.9&\plus{+2.0}&-1.3& \plus{+1.6} \\
    \end{tabular}
    \end{center}
    \caption{Comparison on open-vocabulary semantic segmentation. Baseline results are copied from \cite{xu2023open}.}
    \label{tab:semantic-seg}
    }
    \hfill
    \parbox{.52\textwidth}{
       \begin{center}
        \tablestyle{1.pt}{1.0}
        \begin{tabular}{p{2.6cm}m{1.0cm}m{1.0cm}lll}
         Decoder &  Fusion (things) & Fusion (stuff)  &  \multicolumn{1}{c}{PQ} & \multicolumn{1}{c}{$\text{AP}^{\text{mask}}$} & \multicolumn{1}{c}{oIOU} \\
         \Xhline{0.8pt}
         Unified &  &  & 45.1 & 42.9 & 67.1 \\
         Decoupled & & & 50.6 & 43.6 & 67.6 \\ 
         Unified (\cref{fig:diagram-design-choice}a)  &  \centering 
 \cmark  & \centering \cmark & 44.6 &  42.5 & 66.8 \\
         Decoupled (\cref{fig:diagram-design-choice}b) & \centering \cmark & \centering \cmark  & 50.0 & \bf 44.4 & 77.1 \\ 
         \rowcolor{Gray!60}
         Decoupled (\cref{fig:diagram-design-choice}c) & \centering  \cmark &  & \bf 51.3 & \bf 44.4 & \bf 77.3 \\
         \end{tabular}
         \end{center}
         \caption{An ablation study on different decoder and text-image fusion designs, as depicted in \cref{fig:diagram-design-choice}. 
         We report PQ for panoptic segmentation on MSCOCO, AP$^{\text{mask}}$ for instance segmentation on MSCOCO, and oIoU for referring segmentation on RefCOCO's validation set.
         Our final choice is highlighted in \colorbox{Gray!60}{gray}.}
         \label{tab:ablation_arch}
    }
    \end{table}
}

\def\tabDesignChoice#1{

        \begin{center}
        \begin{tabular}{p{1.5cm}m{1.0cm}m{1.0cm}lll}
         Decoder &  Fusion (things) & Fusion (stuff)  &  \multicolumn{1}{c}{PQ} & \multicolumn{1}{c}{$\text{AP}^{\text{mask}}$} & \multicolumn{1}{c}{oIOU} \\
         \Xhline{0.8pt}
         Unified &  &  & 45.1 & 42.9 & 67.1 \\
         Decoupled & & & 50.6 ( & 43.6 & 67.6 \\ 
         Unified (\cref{fig:diagram-design-choice}a)  &  \centering 
 \cmark  & \centering \cmark & 44.6 &  42.5 (-0.4) & 66.8 \\
         Decoupled (\cref{fig:diagram-design-choice}b) & \centering \cmark & \centering \cmark  & 50.0 (\plus{+4.9}) & 44.4  & 77.1\\ 
         \rowcolor{Gray!60}
         Decoupled (\cref{fig:diagram-design-choice}c) & \centering  \cmark &  & 51.3 (\plus{+6.2}) & 44.4  & 77.3 \\
         \end{tabular}
         \end{center}
         \caption{An ablation study on different decoder and text-image fusion designs, as depicted in \cref{fig:diagram-design-choice}. 
         We report PQ for panoptic segmentation on MSCOCO, AP$^{\text{mask}}$ for instance segmentation on MSCOCO, and oIoU for referring segmentation on RefCOCO's validation set.
         Our final choice is highlighted in \colorbox{Gray!60}{gray}.}
         \label{tab:ablation_arch}
}

\section{Experiments}
\label{sec:experiments}
We comprehensively evaluate \ours through quantitative and qualitative analyses to demonstrate its effectiveness in performing various types of open-vocabulary segmentation and detection tasks.
The implementation details of \ours are explained in \cref{sec:implementation}.
\cref{sec:results} presents the evaluation results of \ours. Additionally, we conduct an ablation study of various design choices in \cref{sec:ablation}.

\subsection{Implementation Details}
\label{sec:implementation}
\noindent \textbf{Model Learning Settings} can be found in our appendix materials. 

\begin{figure}[t]
    \centering
    \includegraphics[width=400px]{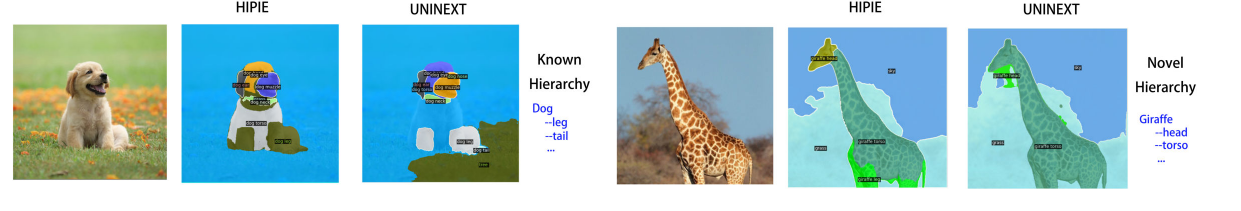}\vspace{-5pt}
    \caption{\textbf{Qualitative Analysis of Open Vocabulary Hierarchal Segmentation}. Because of our hierarchal design, our model produces better-quality masks. In particular, our model can generalize to novel hierarchies that do not exist in part segmentation datasets. 
    }
    \label{fig:hierarchal-demo}
\end{figure}

\noindent \textbf{Evaluation Metrics.}
\textbf{\textit{Semantic Segmentation}} performance is evaluated using the mean Intersection-Over-Union (mIoU) metric. 
For \textbf{\textit{Part segmentation,}} we report mIoU$_\text{PartS}$, which is the mean IoU for part segmentation on grouped part classes \cite{de2021part}.
\textbf{\textit{Object Detection and Instance Segmentation}} results are measured using the COCO-style evaluation metric - mean average precision (AP) \cite{lin2014microsoft}.
\textbf{\textit{Panoptic Segmentation}} is evaluated using the Panoptic Quality (PQ) metric~\cite{kirillov2019panoptic}. 
%
\textbf{\textit{Referring Image Segmentation}} (RIS)~\cite{hu2016segmentation,yu2018mattnet} is evaluated with overall IoU (oIoU).

\subsection{Results}
\label{sec:results}

\tabPanopticSeg{t}
\tabPanopticSegPoster{ht}
\tabOpenSeg{t}
\tabSemanticSeg{t}

\noindent \textbf{Panoptic Segmentation.}
We examine Panoptic Quality (PQ) performance across MSCOCO~\cite{lin2014microsoft} for closed-set and ADE20K~\cite{zhou2019semantic} for open-set zero shot transfer learning. Based on \cref{tab:panoptic-seg} our model is able to outperform the previous close-set state-of-the-art using a ViT-H backbone by +1.8. In addition, we match the best open-set PQ results, while being able to run on more tasks and having a simpler backbone than ODISE \cite{xu2023open}. 
\noindent \textbf{Semantic Segmentation.} 
The evaluation of our model's performance on various open-vocabulary semantic segmentation datasets is presented in \cref{tab:semantic-seg}. These datasets include:
1) A-150: This dataset comprises 150 common classes from ADE20K~\cite{zhou2019semantic}.
2) A-847: This dataset includes all 847 classes from ADE20K~\cite{zhou2019semantic}.
3) PC-59: It consists of 59 common classes from Pascal Context~\cite{mottaghi2014role}.
4) PC-459: This dataset encompasses the full 459 classes of Pascal Context~\cite{mottaghi2014role}.
5) PAS-21: The vanilla Pascal VOC dataset~\cite{everingham2010pascal}, containing 20 foreground classes and 1 background class.
These diverse datasets enable a comprehensive evaluation of our model's performance across different settings, such as varying class sizes and dataset complexities. \cref{tab:semantic-seg} provides insights into how our model performs in handling open-vocabulary semantic segmentation tasks, demonstrating its effectiveness and versatility in detecting and segmenting a wide range of object categories in real-world scenarios.

\begin{figure*}
  \centering
  \includegraphics[width=1.0\textwidth]{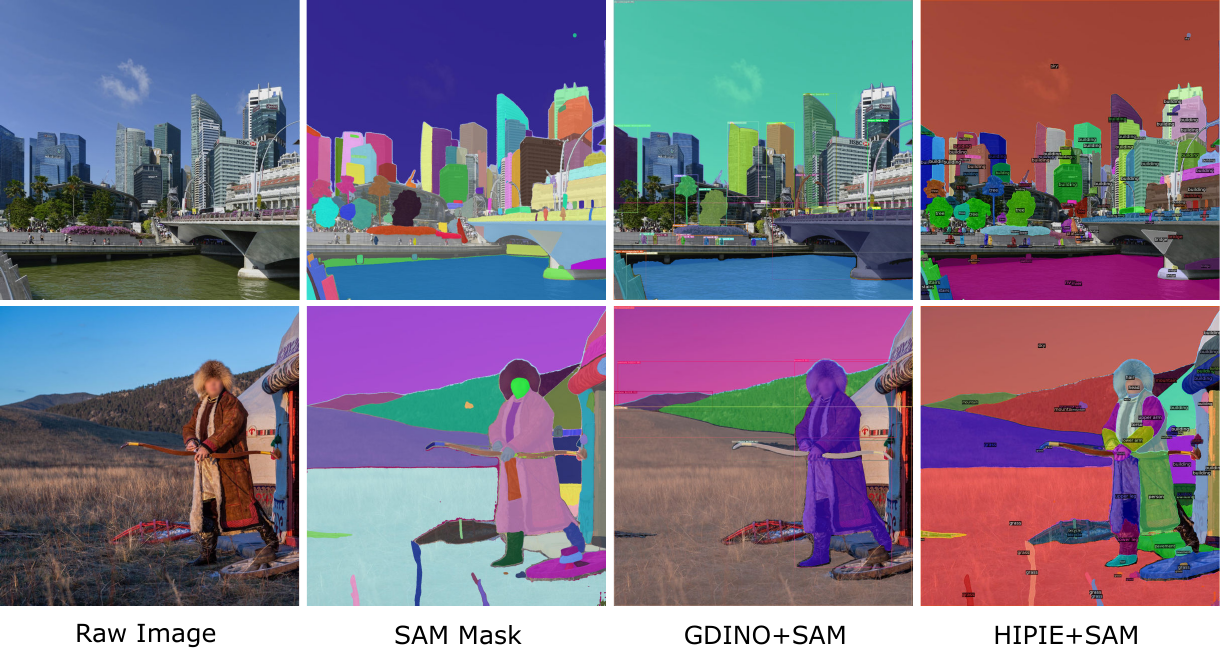}\vspace{-3pt}
  \caption{
   Results of merging \ours with SAM for class-aware image segmentation on SA-1B dataset. 
   Grounded-SAM (Grounding DINO + SAM)~\cite{li2021grounded,kirillov2023segment} cannot fully leverage the benefits of SAM in producing accurate and fine-grained part segmentation masks.
   Our method demonstrates fewer misclassifications and overlooked masks across the SA-1B dataset compared to the Grounded-SAM approach. 
  }
  \label{fig:demo}
\end{figure*}

\noindent \textbf{Part Segmentation.}
We evaluate our models performance on Pascal-Panoptic-Parts dataset \cite{de2021part} and report mIoU$_\text{partS}$ in \cref{tab:panoptic-seg}. We followed the standard grouping from \cite{de2021part}. Our model outperforms state-of-the-art by +5.2 in this metric.
We also provide qualitative comparisons with Grounding DINO + SAM in \cref{fig:demo}. Our findings reveal that the results of Grounding SAM are heavily constrained by the detection performance of Grounding DINO. As a result, they are unable to fully leverage the benefits of SAM in producing accurate and fine-grained part segmentation masks.

\tabDetInstRef{t!}

\noindent \textbf{Object Detection and Instance Segmentation.} We evaluate our model's object detection and instance segmentation capabilities following previous works~\cite{li2022mask,zou2023segment,xu2023open}. On MSCOCO~\cite{lin2014microsoft} and ADE20K~\cite{zhou2019semantic} datasets, \ours achieves an increase of +5.1 and +0.6 AP$^{\text{mask}}$ respectively. Detailed comparisons are provided in \cref{tab:det-inst-seg} which demonstrate state-of-the-art results on ResNet and ViT architectures consistently across all Average Precision metrics.

\noindent \textbf{Referring Segmentation.}
Referring image segmentation (RIS) tasks are examined using the RefCOCO, RefCOCO+, and RefCOCOg datasets. Our model outperforms all the other alternatives by an average of +0.5 in overall IoU (oIoU).

\subsection{Ablation Study}
\label{sec:ablation}

To demonstrate the effectiveness of our design choices for text-image fusion mechanisms and representation learning modules for stuff and thing classes, we conduct an ablation study (depicted in \cref{fig:diagram-design-choice}) and present the results in \cref{tab:ablation_arch}. From this study, we draw several important conclusions:
\textit{\textbf{1)}} Text-image fusion plays a critical role in achieving accurate referring segmentation results.
\textit{\textbf{2)}} The early text-image fusion approach for stuff classes negatively impacts the model's performance on panoptic segmentation. This finding validates our analysis in the introduction section, where we highlighted the challenges introduced by the high levels of confusion in stuff's textual features, which can adversely affect the quality of representation learning.
\textit{\textbf{3)}} Our design choices significantly improve the performance of panoptic segmentation, instance segmentation, and referring segmentation tasks.
These conclusions underscore the importance of our proposed design choices in achieving improved results across multiple segmentation tasks.

%% file: sections/5_conclusion.tex
\section{Conclusions}
\label{sec:conclusion}
This paper presents \ours, an open-vocabulary, universal, and hierarchical image segmentation model that is capable of performing various detection and segmentation tasks using a unified framework, inculding object detection, instance-, semantic-, panoptic-, hierarchical-(whole instance, part, subpart), and referring-segmentation tasks. 
Our key insight is that we should decouple the representation learning modules and text-image fusion mechanisms for background (\ie, referred to as stuff) and foreground (\ie, referred to as things) classes.
Extensive experiments demonstrate that \ours achieves state-of-the-art performance on diverse datasets, spanning across a wide range of tasks and segmentation granularity. 

\textbf{Acknowledgement}
Trevor Darrell and XuDong Wang were funded by DoD including DARPA LwLL and the Berkeley AI Research (BAIR) Commons.

%% file: sections/6_appendix.tex
\def\figDemoAppendix#1{
    \captionsetup[sub]{font=small}
    \begin{figure}[#1]
      \centering
      \includegraphics[width=1.0\textwidth]{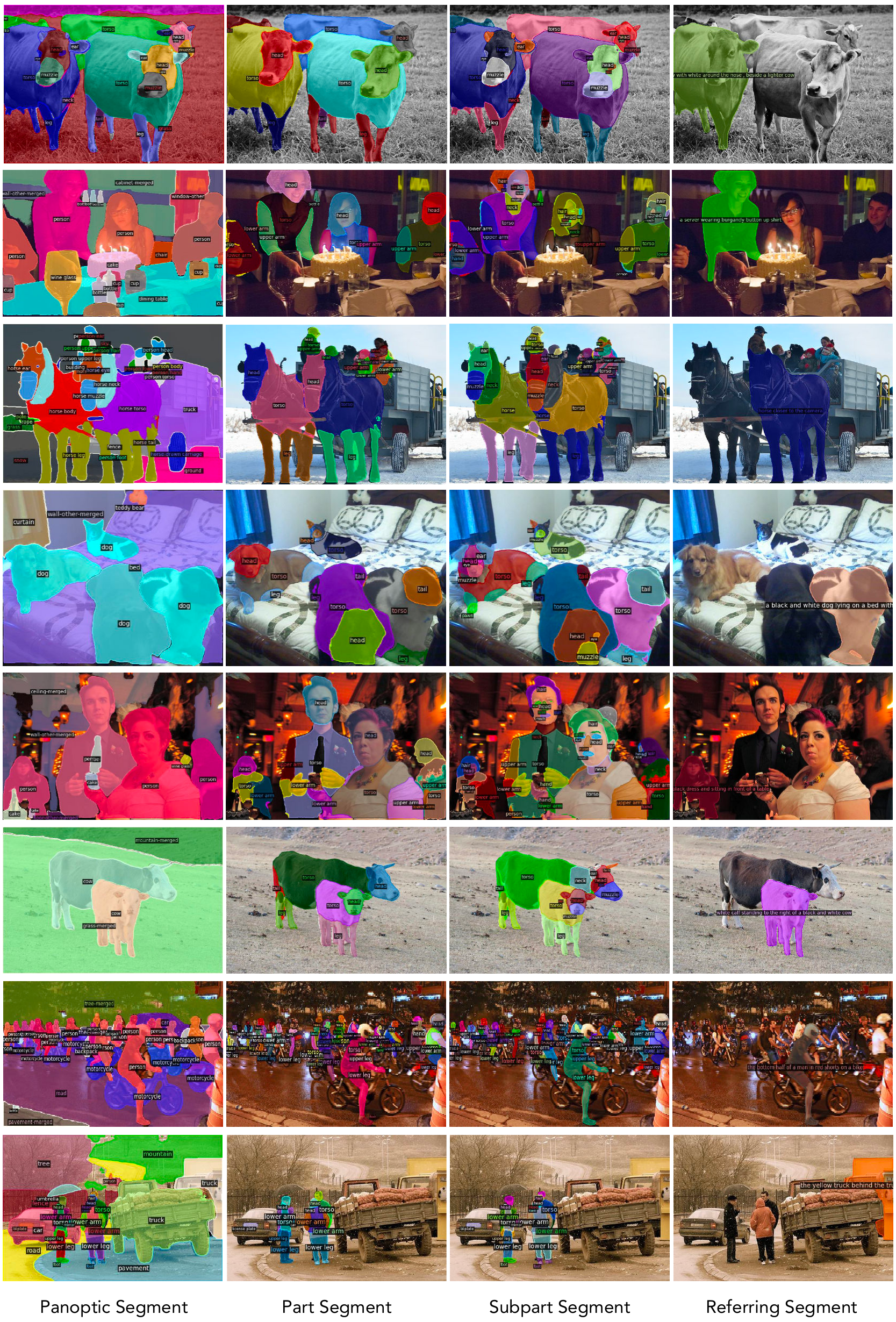}
      \caption{
      More visualizations showcasing panoptic segmentation, part segmentation, subpart segmentation, and referring segmentation results on RefCOCO. It is recommended to view the results in color and zoom in for better detail.
      }
      \label{fig:demo-appendix}
    \end{figure}
}

\def\tabSeginW#1{
\begin{table}[#1]
\tablestyle{1.2pt}{1.1}
    \begin{center}
    \tiny
    \begin{tabular}{l|cc|ccccccccccccccccccccccccc}
        Method & \rotatebox{0}{\bf Mean} & \rotatebox{0}{\bf Median} & \rotatebox{90}{\tablestyle{0.8pt}{0.6}\tiny\begin{tabular}[c]{@{}l@{}}Airplane \\Parts\end{tabular}} & \rotatebox{90}{Bottles} & \rotatebox{90}{\tablestyle{0.8pt}{0.6}\tiny\begin{tabular}[c]{@{}l@{}}Brain \\Tumor\end{tabular}} & \rotatebox{90}{Chicken} & \rotatebox{90}{Cows} & \rotatebox{90}{\tablestyle{0.8pt}{0.6}\tiny\begin{tabular}[c]{@{}l@{}}Electric\\Shaver\end{tabular}} & \rotatebox{90}{Elephants} & \rotatebox{90}{Fruits} & \rotatebox{90}{Garbage} & \rotatebox{90}{\tablestyle{0.8pt}{0.6}\tiny\begin{tabular}[c]{@{}l@{}}Ginger\\Garlic\end{tabular}} & \rotatebox{90}{Hand} & \rotatebox{90}{\tablestyle{0.8pt}{0.6}\tiny\begin{tabular}[c]{@{}l@{}}Hand\\Metal\end{tabular}} & \rotatebox{90}{\tablestyle{0.8pt}{0.6}\tiny\begin{tabular}[c]{@{}l@{}}House\\Parts\end{tabular}} & \rotatebox{90}{\tablestyle{0.8pt}{0.6}\tiny\begin{tabular}[c]{@{}l@{}}HouseHold\\Items\end{tabular}} & \rotatebox{90}{\tablestyle{0.8pt}{0.6}\tiny\begin{tabular}[c]{@{}l@{}}Nutterfly\\Squireel\end{tabular}} & \rotatebox{90}{Phones} & \rotatebox{90}{Poles} & \rotatebox{90}{Puppies} & \rotatebox{90}{Rail} & \rotatebox{90}{\tablestyle{0.8pt}{0.6}\tiny\begin{tabular}[c]{@{}l@{}}Salmon\\Fillet\end{tabular}} & \rotatebox{90}{Strawberry} & \rotatebox{90}{Tablets}  & \rotatebox{90}{Toolkits}  & \rotatebox{90}{Trash}  & \rotatebox{90}{Watermelon} \\
        \Xhline{0.8pt}
        X-Decoder(L) & 32.3 & 22.3 & 13.1 & 42.1 & \textbf{2.2} & 8.6 & 44.9 & 7.5 & 66.0 & \textbf{79.2} & \textbf{33.0} & 11.6 & 75.9 & 42.1 & \textbf{7.0} & 53.0 & 68.4 & 15.6 & 20.1 & 59.0 & \textbf{2.3} & 19.0 & 67.1 & \textbf{22.5} & 9.9 & 22.3 & 13.8 \\
        \ours(H) & \textbf{41.2} & \textbf{45.1} & \textbf{14.0} & \textbf{45.1} & 1.9 & \textbf{46.5} & \textbf{50.1} & \textbf{76.1} & \textbf{68.6} & 61.1 & 31.2 & \textbf{24.3} & \textbf{94.2} & \textbf{64.0} & 6.8 & \textbf{53.4} & \textbf{79.7} & 7.0 & 6.7 & \textbf{64.6} & 2.2 & \textbf{41.8} & \textbf{81.5} & 8.8 & \textbf{17.9} & \textbf{31.2} & \textbf{50.6} 
    \end{tabular}
    \end{center}
\caption{Segmentation Result on SeginW benchmark across 25 datasets. We report mAP. We outperform X-Decoder by a large margin (+8.9)}
\label{tab:seginW}
\end{table}
}

\def\tabOdinW#1{
\begin{table}[#1]
\tablestyle{0.2pt}{1.1}
    \tiny
    \begin{tabular}{l|x{0.4cm}x{0.4cm}|ccccccccccccccccccccccccccccccccccc}
    Method & \rotatebox{90}{\textbf{Mean}} & \rotatebox{90}{\textbf{Median}} & \rotatebox{90}{\tablestyle{0.8pt}{0.6}\tiny\begin{tabular}[c]{@{}l@{}}AerialMaritime\\ Drone large\end{tabular}} & \rotatebox{90}{\tablestyle{0.8pt}{0.6}\tiny\begin{tabular}[c]{@{}l@{}}AerialMaritime\\Drone tiled\end{tabular}} & \rotatebox{90}{\tablestyle{0.8pt}{0.6}\tiny\begin{tabular}[c]{@{}l@{}}American Sign\\Lang Letters\end{tabular}} & \rotatebox{90}{Aquarium} & \rotatebox{90}{BCCD} & \rotatebox{90}{boggleBoards} & \rotatebox{90}{\tablestyle{0.8pt}{0.6}\tiny\begin{tabular}[c]{@{}l@{}}brackish\\Underwater\end{tabular}} & \rotatebox{90}{ChessPieces} & \rotatebox{90}{\tablestyle{0.8pt}{0.6}\tiny\begin{tabular}[c]{@{}l@{}}Cottontail\\Rabbits\end{tabular}} & \rotatebox{90}{\tablestyle{0.8pt}{0.6}\tiny\begin{tabular}[c]{@{}l@{}}dice\\mediumColor\end{tabular}} & \rotatebox{90}{DroneControl} & \rotatebox{90}{\tablestyle{0.8pt}{0.6}\tiny\begin{tabular}[c]{@{}l@{}}EgoHands\\generic\end{tabular}} & \rotatebox{90}{\tablestyle{0.8pt}{0.6}\tiny\begin{tabular}[c]{@{}l@{}}EgoHands\\specific\end{tabular}} & \rotatebox{90}{\tablestyle{0.8pt}{0.6}\tiny\begin{tabular}[c]{@{}l@{}}HardHat \\Workers\end{tabular}} & \rotatebox{90}{Mask Wearing} & \rotatebox{90}{\tablestyle{0.8pt}{0.6}\tiny\begin{tabular}[c]{@{}l@{}}Mountain Dew\\Commercial\end{tabular}} & \rotatebox{90}{\tablestyle{0.8pt}{0.6}\tiny\begin{tabular}[c]{@{}l@{}}North America\\Mushrooms\end{tabular}} & \rotatebox{90}{\tablestyle{0.8pt}{0.6}\tiny\begin{tabular}[c]{@{}l@{}}open Poetry\\Vision\end{tabular}} & \rotatebox{90}{\tablestyle{0.8pt}{0.6}\tiny\begin{tabular}[c]{@{}l@{}}Oxford Pets\\by-breed\end{tabular}} & \rotatebox{90}{\tablestyle{0.8pt}{0.6}\tiny\begin{tabular}[c]{@{}l@{}}Oxford Pets\\by-species\end{tabular}} & \rotatebox{90}{Packages} & \rotatebox{90}{Pascal VOC} & \rotatebox{90}{Pistols} & \rotatebox{90}{PKLot} & \rotatebox{90}{plantdoc} & \rotatebox{90}{Pothole} & \rotatebox{90}{Raccoon} & \rotatebox{90}{selfdriving Car} & \rotatebox{90}{\tablestyle{0.8pt}{0.6}\tiny\begin{tabular}[c]{@{}l@{}}Shellfish\\OpenImages\end{tabular}} & \rotatebox{90}{\tablestyle{0.8pt}{0.6}\tiny\begin{tabular}[c]{@{}l@{}}Thermal \\Cheetah\end{tabular}} & \rotatebox{90}{\tablestyle{0.8pt}{0.6}\tiny\begin{tabular}[c]{@{}l@{}}thermal Dogs\\And People\end{tabular}} & \rotatebox{90}{Uno Cards} & \rotatebox{90}{\tablestyle{0.8pt}{0.6}\tiny\begin{tabular}[c]{@{}l@{}}Vehicles\\OpenImages\end{tabular}}  & \rotatebox{90}{\tablestyle{0.8pt}{0.6}\tiny\begin{tabular}[c]{@{}l@{}}website\\Screenshots\end{tabular}}  & \rotatebox{90}{Wildfire Smoke} \\
    \Xhline{0.8pt}
    MDETR & 10.7 & 3.0 & 0.6 & 5.4 & 0.3 & 1.7 & 6.7 & 0.0 & 0.7 & 3.0 & 66.5 & 0.0 & 3.8 & 5.9 & \textbf{3.5} & 0.4 & 0.4 & 3.0 & 39.8 & 0.0 & 0.0 & 0.7 & 63.6 & 5.6 & 15.9 & 0.0 & 0.5 & \textbf{12.7} & \textbf{50.6} & 2.8 & 8.1 & \textbf{4.5} & 42.8 & 0.0 & 13.4 & \textbf{0.7} & 12.5 \\
    GLIP-T & 11.4 & 1.6 & 8.3 & \textbf{17.1} & 0.1 & 16.0 & 1.7 & 0.0 & 1.7 & 0.0 & 57.0 & 0.5 & 0.1 & 1.1 & 0.1 & \textbf{2.7} & 0.6 & 15.3 & 5.9 & 0.0 & 0.3 & 1.6 & 58.3 & 51.2 & 31.6  & 0.0 & 1.6 & 1.6 & 6.2 & \textbf{7.4} & 15.9 & 0.2 & 38.7 & 0.0 & \textbf{55.0} & 0.3 & 0.0 \\
    \ours$^{\dagger}$ & 14.5 & {3.9} & {5.2} & 9.6 & \textbf{2.9} & 8.6 & 6.0 & 0.0 & 0.9 & 3.8 & 69.5 & \textbf{0.5} & 0.7 & 5.8 & 0.2 & 1.4 & 0.8 & \textbf{37.7} & 27.4 & 0.0 & \textbf{7.8} & 2.5 & \textbf{68.1} & 58.6 & 36.4 & 1.1 & \textbf{3.7} & 3.9 & 33.4 & 5.3 & 27.5 & 0.5 & 24.5 & 0.0 & 53.9 & 0.3 & 0.0 \\
    \ours$^{\ddagger}$ & \textbf{17.9} & \textbf{5.5} & \textbf{10.9} & 16.6 & 2.8 & \textbf{18.3} & \textbf{8.0} & \textbf{0.1} & \textbf{2.7} & \textbf{5.5} & \textbf{75.7} & 0.3 & \textbf{1.6} & \textbf{6.6} & 0.5 & 1.8 & \textbf{1.1} & 8.5 & \textbf{42.7} & 0.0 & 7.2 & \textbf{2.7} & 56.2 & \textbf{66.0} & \textbf{66.8} & \textbf{2.6} & 3.6 & 2.9 & 49.7 & 7.3 & \textbf{49.6} & 0.3 & \textbf{53.3} & 0.0 & 53.5 & 0.4 & 0.3
    \end{tabular}
    \vspace{2pt}
    \caption{
    We present the object detection results in the OdinW benchmark. We report mAP and mean results averaged over 35 datasets. Notably, our ResNet-50 baseline surpasses GLIP-T by $+3.1$. We use the notation \ours$^{\dagger}$ and \ours$^{\ddagger}$ to denote our method with ResNet-50 and ViT-H backbones, respectively.}
    \label{tab:odinw}
\end{table}
}

\def\figDemoAppendixTwo#1{
    \captionsetup[sub]{font=small}
    \begin{figure}[#1]
      \centering
      \includegraphics[width=1.0\textwidth]{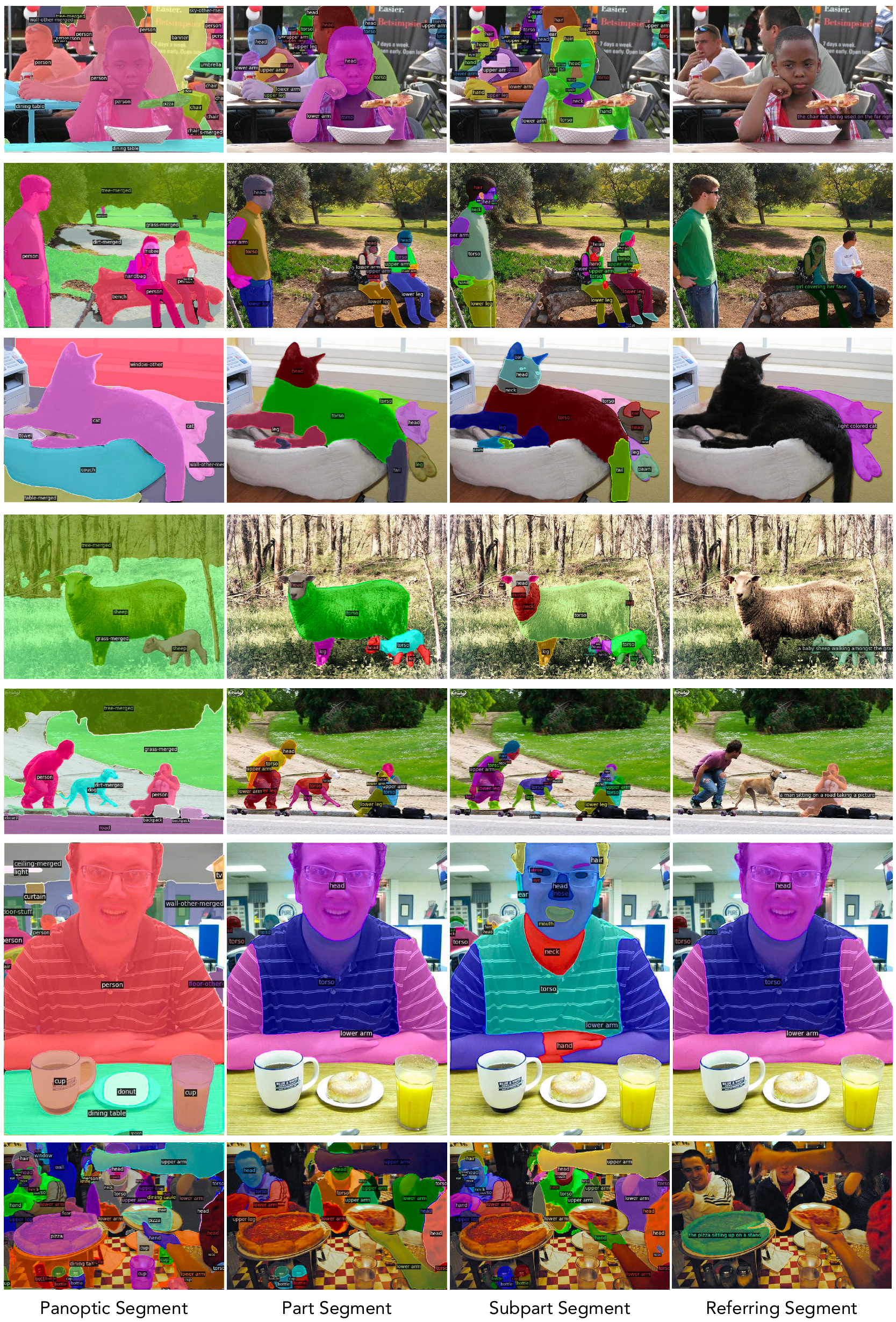}
      \caption{
      More visualizations showcasing panoptic segmentation, part segmentation, subpart segmentation, and referring segmentation results on RefCOCO. It is recommended to view the results in color and zoom in for better detail.
      }
      \label{fig:demo-appendix-two}
    \end{figure}
}

\def\figDemoAppendixThree#1{
    \captionsetup[sub]{font=small}
    \begin{figure}[#1]
      \centering
      \includegraphics[width=1.0\textwidth]{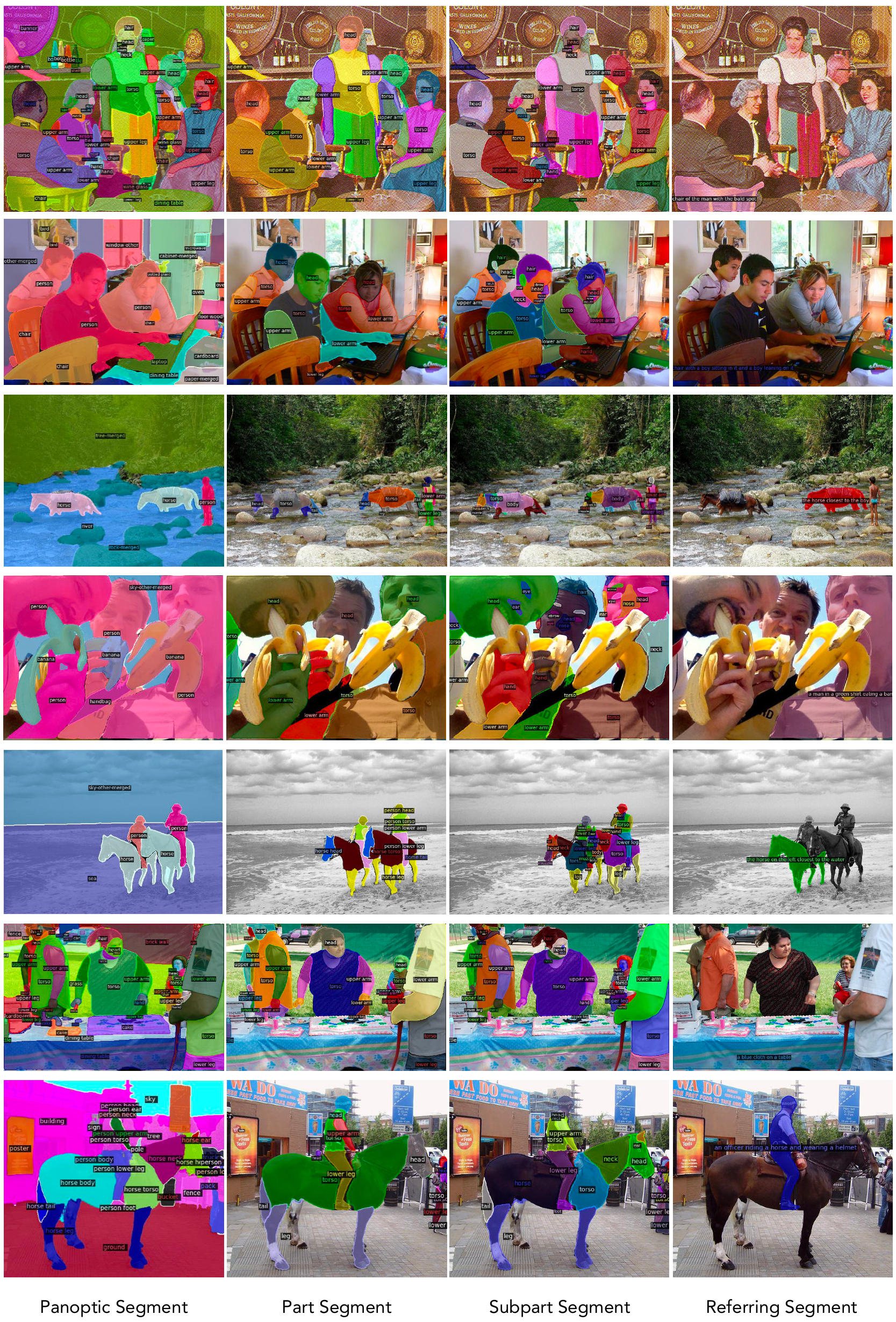}
      \caption{
      More visualizations showcasing panoptic segmentation, part segmentation, subpart segmentation, and referring segmentation results on RefCOCO. It is recommended to view the results in color and zoom in for better detail.
      }
      \label{fig:demo-appendix-three}
    \end{figure}
}

\setcounter{section}{0}
\renewcommand{\thesection}{A.\arabic{section}}
\setcounter{equation}{0}
\renewcommand{\theequation}{A\arabic{equation}}
\setcounter{table}{0}
\renewcommand{\thetable}{A\arabic{table}}
\setcounter{figure}{0}
\renewcommand{\thefigure}{A\arabic{figure}}

\section*{\Large Appendix}
\label{sec:appendix}
\section{List of Datasets}

\begin{table}[ht]
\begin{tabular}{c|ccccc|c|c}
  & semantic                & instance             & panoptic                  & grounding & part                 & training & \# images \\
  \hline
ADE-150               & \checkmark                    & \checkmark                     & \checkmark                     &           &                      &          & 2000       \\
Pascal VOC             & \checkmark                     &                      &                      &           &                      &          & 1449       \\
Pascal Context-59     & \checkmark                     &                      &                      &           &                      &          & 5105       \\
Pascal-Panoptic-Parts & \checkmark                     & \checkmark                     & \checkmark                     &          & \checkmark                     & *        & 10103      \\
COCO                  & \checkmark                     & \checkmark                     & \checkmark                     &        &  & \checkmark         & 121408     \\
RefCOCO               &  &  &  & \checkmark          &  & \checkmark         & 19994     \\
RefCOCO+             &  &  &  & \checkmark          &  & \checkmark         & 19992     \\
RefCOCOg              &  &  &  & \checkmark          &  & \checkmark         & 26711  \\
\end{tabular}
\caption{\textbf{List of the dataset used.} The checkmarks denote whether a dataset has a particular type of annotation and whether the dataset is used in the training process. * Because of a data leak between Pascal-Panoptic-Parts  and other Pascal datasets, we use weights trained without Pascal-Panoptic-Parts in those evaluations unless otherwise specified. }
\label{tab:list_of_dataset}
\end{table}

We report the statistics of datasets used in training and evaluation in table \cref{tab:list_of_dataset}. Additionally, we further evaluate our model on 35 object detection datasets and 25 segmentation datasets in \cref{sec:odinw}. In total, we benchmarked our model on around 70 datasets. These benchmarks show our model can adapt to many different scenarios and retain a reasonable performance in a zero-shot manner.

\section{Hierarchical Segmentation}
\begin{figure}[h]
    \centering
    \includegraphics[width=0.7\textwidth]{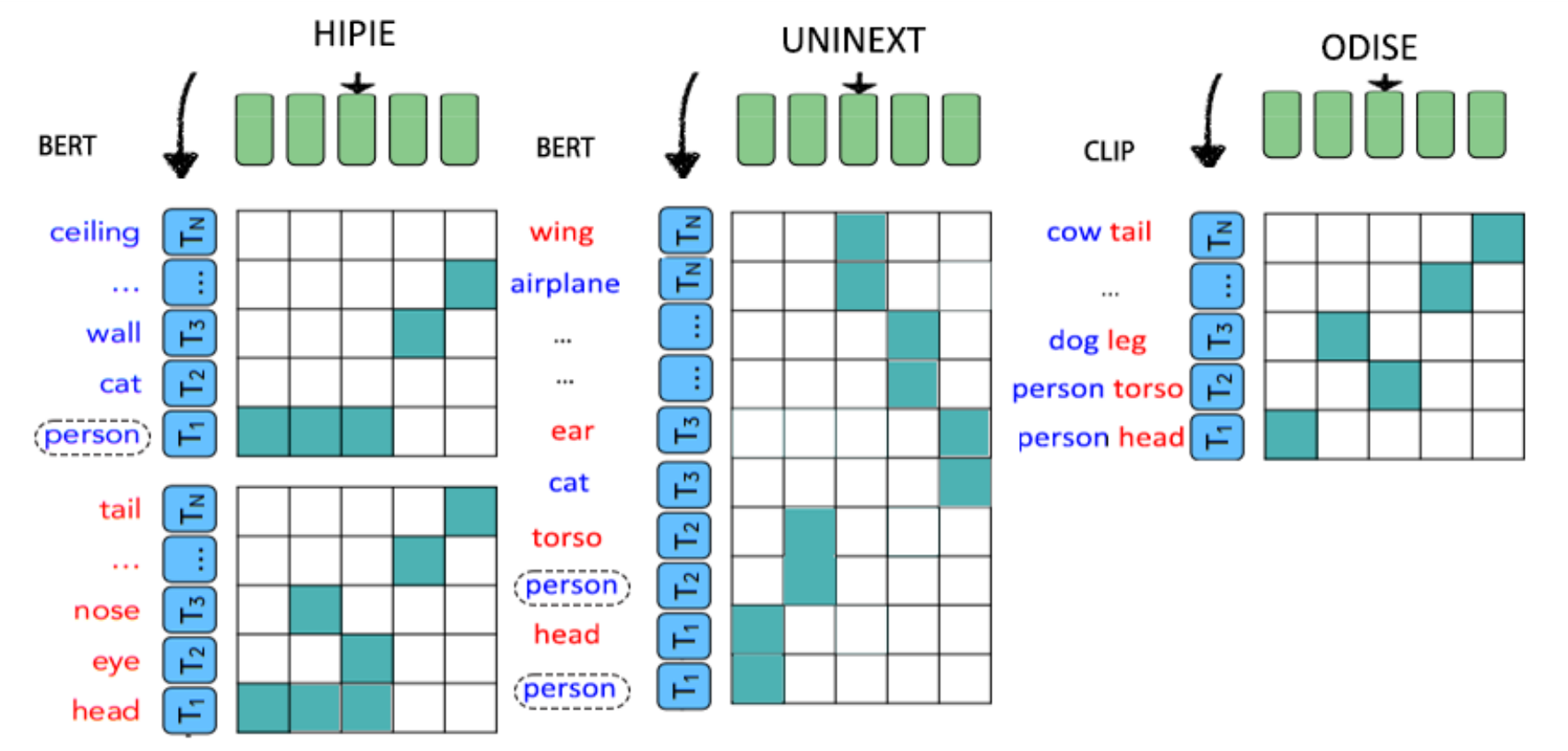}\vspace{-3pt}
    \caption{\textbf{Hierarchal Design of HIPIE compared with other methods}. 
    }
    \label{fig:hierarchal}
\end{figure}

\vspace{3pt}
\cref{fig:hierarchal} highlights the difference of our approach with other methods for hierarchical segmentation. We concatenate class names
of different hierarchies as prompts. During the training, we uniquely contrast a mask embedding with both scene-level and part-level labels explicitly. Previous works such as UNINEXT and ODISE only treat these classes as normal multi-word labels. While UNINEXT allows contrasting different words individually because of the design of BERT encoder, it leads to suboptimal signals. In the example above, "person head" has both positive and negative target for "person".

\section{Experiment Setup}

\subsection{Model Learning Settings}
\ours is first pre-trained on Objects365~\cite{shao2019objects365} for 340k iterations, using a batch size of 64 and a learning rate of 0.0002, and the learning rate is dropped by a factor of 10 after the 90th percentile of the schedule. 
After the pre-training stage, we finetune \ours on COCO~\cite{lin2014microsoft}, RefCOCO, RefCOCOg, and RefCOCO+~\cite{nagaraja2016modeling,yu2016modeling} jointly for 120k iterations, using a batch size of 32 and a learning rate of 0.0002. 
For both stages, we resize the original images so that the shortest side is at least 800 pixels and at most 1024 pixels, while the longest side is at most 1333 pixels. 
For part segmentation, we train additionally train our model jointly on Pascal-Panoptic-Parts \cite{de2021part} dataset and all previously mentioned datasets. Because of potential data leaks between Pascal-Panoptic-Parts and other Pascal datasets used in the open-vocabulary segmentation evaluation, we report those numbers with weights not trained on Pascal-Panoptic-Part dataset.  Because of our hierarchal design, our model produces better-quality masks. In particular, our model can generalize to novel hierarchies that do not exist in part segmentation datasets.  In \cref{fig:hierarchal-demo}, we provide visualization of such results.

\subsection{Implementation Details}
For loss functions in \cref{eq:loss}, we have $\lambda_{cls}=2.0,\lambda_{mask}=5.0,\lambda_{box}=5.0,\lambda_{ce}=1.0,\lambda_{dice}=1.0,\lambda_{L1}=1.0,\lambda_{giou}=0.2$. For $\lambda$ in \cref{eq:open_vocab}, we use $\lambda=0.2$ for seen classes during the training and $\lambda=0.45$ for novel classes. In close-set evaluation, we set $\lambda=0.0$ and do not use CLIP. We also do not use CLIP for PAS-21 evaluation (whose classes are mostly covered by COCO) because we find it degrades the performance. We use 800 and 1024-resolution images during the training. For evaluations, we use 1024-resolution images.
\subsection{Training Process}
\begin{table}[ht]
    \centering
\begin{tabular}{c|c|cHcHHHHcc}

\toprule
Stage & Task & Dataset &  Weight & Batch Size & Short & Long & Num GPU & Lr & Max Iter & Step \\

\midrule
\multirow{1}{*} {\uppercase\expandafter{\romannumeral1}}& \multirow{1}{*} {OD\&IS} & Objects365~ & 1 &  64 & $480\sim800$ & 1333 & 32 & 0.0002 & 340741 & 312346 \\
\midrule
\multirow{2}{*}{\uppercase\expandafter{\romannumeral2}} & OD\&IS & COCO~ & 1 & 32 & $480\sim800$ & 1333 & \multirow{2}{*}{16} & \multirow{2}{*}{0.0002} & \multirow{2}{*}{91990} & \multirow{2}{*}{76658} \\
& REC\&RIS & RefCOCO/g/+~ & 1 & 32 & $480\sim800$ & 1333 & & & & \\
\midrule
\multirow{3}{*}{\uppercase\expandafter{\romannumeral3}} & PanoS & COCO~ & 1 & 32 & $480\sim800$ & 1333 & \multirow{3}{*}{16} & \multirow{3}{*}{0.0002} & \multirow{3}{*}{150000} & \multirow{3}{*}{100000,135000} \\
& REC\&RIS & RefCOCO/g/+~ & 1 & 32 & $480\sim800$ & 1333 & & & & \\
& PartS & Pascal-Panoptic-Parts & 1 & 32 & $480\sim800$ & 1333 & & & & \\

\bottomrule
\end{tabular}
\vspace{5pt}\caption{\textbf{Training Process.} Following UNINEXT \cite{yan2023universal}, We first pretrain our model for object detection on Object365 for 340k iteration (Stage I). Then we fine-tune our model jointly on COCO for object detection, instance segmentation, referring expression comprehension (REC), and referring segmentation (RIS) for 92k iteration (Stage II). We further jointly train our model on Panoptic Segmentation, REC, RIS, and Part Segmentation for 150k iteration (Stage III)}
\label{tab:training_process}
\end{table}

We train all our models on NVIDIA-A100 GPUs with a batch size of 2 per GPU using AdamW 
\cite{loshchilovdecoupled} optimizer. We use a base learning rate of 0.0001 and a weight decay of 0.01. The learning rate of the backbone is further multiplied by 0.1.  Following UNINEXT \cite{yan2023universal}, We first pretrain our model for object detection on Object365 for 340k iteration (Stage I). Then we fine-tune our model jointly on COCO for object detection, instance segmentation, referring expression comprehension (REC), and referring segmentation (RIS) for 91k iteration (Stage II). We further jointly train our model on Panoptic Segmentation, REC, RIS, and Part Segmentation for 150k iteration (Stage III).  In Stage I, the learning rate is dropped by a factor of 10 after 312k iterations. In stage II, the learning rate is dropped by a factor of 10 after 77k iterations. In Stage III, the learning rate is dropped by a factor of 10 after 100k and 135k iterations. In all stages, we sample uniformly across datasets when there are multiple datasets. The global batch size is 64 in Stage I and 32 in Stage II and III. Notably, our stage I and II is identical to the setup of UNINEXT. For ablation studies, we train stage III only and reduce the schedule to 90k iterations. The learning rate schedule is also scaled accordingly. The details of training recipe is shown in \cref{tab:training_process}.
\section{Additional Evaluations}
\subsection{Referring Expression Comprehension }

\begin{table}[ht]
\begin{center}
    \begin{tabular}{p{1.9cm}lHHlHcclHcclHcc}
    \multirow{2}{*}{Method} & \multirow{2}{*}{Backbone} & \multirow{2}{*}{\#Params} & \multirow{2}{*}{Data} && \multicolumn{3}{c}{{RefCOCO}} && \multicolumn{3}{c}{{RefCOCO+}} && \multicolumn{3}{c}{{RefCOCOg}} \\
    \cline{6-8} \cline{10-12} \cline{14-16}
    &&&&& {} & {oIoU} & P@0.5 && {} & {oIoU} &P@0.5 && {} & {oIoU} & P@0.5\\ [.1em]
    \Xhline{0.8pt}
    MAttNet~\cite{yu2018mattnet} & RN101 & - &&& \phantom{1}\phantom{1}- &56.5 & 76.7 && \phantom{1}\phantom{1}- & 46.7 & 65.3 && \phantom{1}\phantom{1}- & 47.6 & 66.6 \\
    VLT~\cite{ding2022vlt} & Dark56 & - &&& \phantom{1}\phantom{1}- & 65.7 & 76.2 && \phantom{1}\phantom{1}- & 55.5 & 64.2 && \phantom{1}\phantom{1}- & 53.0 & 61.0 \\
    RefTR~\cite{muchen2021referring} & RN101 & - &&& \phantom{1}\phantom{1}- & 74.3 & 85.7 && \phantom{1}\phantom{1}- & 66.8 & 77.6 && \phantom{1}\phantom{1}- & 64.7 & 82.7 \\
    UNINEXT~\cite{yan2023universal} & RN50 & - &&& \phantom{1}\phantom{1}- & 77.9 & 89.7 && \phantom{1}\phantom{1}- & 66.2 & 79.7 && \phantom{1}\phantom{1}- & 70.0 & 84.0 \\
    UNINEXT~\cite{yan2023universal} & ViT-H & - &&& \phantom{1}\phantom{1}- & 82.2 & 92.6 && \phantom{1}\phantom{1}- & 72.5 & 85.2 && \phantom{1}\phantom{1}- & 74.7 & 88.7 \\
    \rowcolor{Gray!60}
    \ours & RN50 & TBD &&   && 78.3 &   90.1 & && 66.2 & 80.0 & & & 69.8  &83.6  \\
    \rowcolor{Gray!60}
    \ours & ViT-H & TBD && & &82.6& 93.0&& &73.0&85.5 & &&75.3&88.9 \\
    \rowcolor{Gray!60}
    \multicolumn{2}{l}{\textit{vs. prev. SOTA}} &&& & \plus{+0.4} & \plus{+0.4} & \plus{+0.4} & & & \plus{+0.5} &\plus{+0.3} & & \plus{} & \plus{+0.6} & \plus{+0.2} \\
    \end{tabular}
    \end{center}
    \caption{Comparison on the referring expression comprehension (REC), and referring image segmentation (RIS) tasks. The evaluation is carried out on the validation sets of RefCOCO, RefCOCO+, and RefCOCOg datasets using Precision@0.5 and overall IoU (oIoU) metrics for REC and RIS, respectively.}
    \label{tab:refer-comp}
\end{table}

In addition to Referring Segmentation reported in \cref{tab:refer-seg}, 
we further report results on Referring Expression Comprehension (REC), which aims to locate a target object in an image at the pixel-level, given a referring expression as input. 
We establish new state-of-the-art performance by an average of $+0.3$ P@0.5 and $+0.5$ oIoU across three datasets.

\tabOdinW{t!}
\tabSeginW{t!}

\subsection{Object Detection and Segmentation in the Wild}
\label{sec:odinw}
To further examine the open-vocabulary capability of our model, we evaluate it on the Segmentation in the Wild (SeginW) \cite{zou2022generalized} consisting of 25 diverse segmentation datasets and Object Detection in the Wild (OdinW) \cite{li2021grounded} Benchmark consisting of 35 diverse detection datasets. Since OdinW benchmark contains Pascal VOC and some of the classes in SeginW benchmark are covered by Pascal-Panoptic-Parts, we use a version of our model that is not trained on Pascal-Panoptic-Parts for both benchmarks for a fair comparison.

We report the results in \cref{tab:seginW} and \cref{tab:odinw}. Notably, our method establishes a new state-of-the-art of SeginW benchmark by an average of $+8.9$ mAP across 25 datasets.  We achieve comparable performance under similar settings. In particular, our ResNet-50 baseline outperforms GLIP-T by $+3.1$ mAP. We note that other methods such as GroundingDINO \cite{liu2023grounding} achieve better absolute performance by introducing more grounding data, which can be critical in datasets whose classes are not common objects. (For example, the classes of  Boggle Boards are letters, the classes of UnoCards are numbers, and the classes of websiteScreenshots are UI elements). 

\section{Other Ablation Studies}
\label{sec:other_ablations}
\begin{table}[ht]
    \centering
    \begin{tabular}{c|cc|c}
     & \multicolumn{2}{c|}{COCO} & RefCOCO\\
               & PQ & \phantom{2}AP$^{\text{Mask}}$ & oIoU \\
               \hline
       CLIP  &  \textbf{51.5} & 44.3 & 48.7\\
       BERT  &  51.3  & \textbf{44.4} & \textbf{77.3}
    \end{tabular}
    \caption{Ablation Studies on the choice of Text Encoder. We find that while CLIP and BERT achieve similar performance on panoptic and instance segmentation, BERT performs significantly better on Referring Instance Segmentation ($+28.6$ oIoU).} 
    \label{tab:ablation_text_encoder}
\end{table}

\begin{table}[ht]
    \centering
    \begin{tabular}{c|cc|c}
     & \multicolumn{2}{c|}{COCO} & RefCOCO\\
               & PQ & \phantom{2}AP$^{\text{Mask}}$ & oIoU \\
               \hline
       w/o OTA  &  50.9 & 43.6 & 76.3\\
       w/ OTA  &  \textbf{51.3 } & \textbf{44.4} & \textbf{77.3}
    \end{tabular}
    \caption{Ablation Studies on the SimOTA matching process. Introducing SimOTA leads to performance improvement in all evaluation metrics. }
    \label{tab:ablation_ota}
\end{table}
We provide further ablations on a few design choices in this section. 

\textbf{Text Encoder}. We experiment with replacing the BERT text encoder in UNINEXT with a pre-trained CLIP encoder. Additionally, following practices of ODISE \cite{xu2023open}, we prompt each label to a sentence "a photo of <label>". For RIS and REC tasks, the language expression remains unchanged. We report the result in \cref{tab:ablation_text_encoder}. We find that while CLIP and BERT achieve similar performance on panoptic and instance segmentation, BERT performs significantly better on referring instance segmentation ($+28.6$ oIoU). We hypothesize that this may be caused by the lack of explicit language-focused training which can help achieve a better understanding of referring expression.  

\textbf{SimOTA}.Following UNINEXT \cite{yan2023universal} we adopted simOTA in the matching process for "thing" classes during the training. We experiment with removing simOTA matching and use standard one-to-one matching instead. We report the result in \cref{tab:ablation_ota}.  We find that simOTA improves the performance on both panoptic segmentation and referring instance segmentation. 


\section{Limitations}
\label{sec:limitations}
We've showcased experimental evidence supporting our method across a diverse set of tasks, including open vocabulary panoptic and semantic segmentation, instance and referring segmentation, and object detection. However, it will be crucial for future work to test our methodology on video-related tasks, such as object tracking and segmentation, to draw comparisons with other universal models like UNINEXT \cite{yan2023universal}. Furthermore, it's worth considering additional pretraining of our vision encoder on newer, more complex datasets that encompass a vast amount of masks and information. For instance, SA-1B \cite{kirillov2023segment}, which includes over 1 billion masks, would serve as an ideal training ground. Lastly, it would be advantageous to measure the change in performance when training on supplementary hierarchical datasets. Such an approach will allow us to demonstrate more varied object part segmentations, thereby expanding the capabilities and versatility of our model.

\section{Broader Impact}
Our research introduces a potent approach to hierarchical and universal open vocabulary image segmentation, aiming to address the ever-increasing demand for more data and advanced model architectures. As the demand increases, practical methodologies such as universal segmentation are projected to play a vital role in constructing and training foundational models. Our model, \ours, shows promise for fostering progress in a multitude of fields where hierarchical data are critical, including self-driving cars, manufacturing, and medicine. However, it's imperative to acknowledge potential limitations. Given that our model is trained on human annotations and feedback, it can inadvertently replicate any errors or biases present in the datasets. The architecture's complexity is further enhanced when multiple models are integrated, which, in turn, compromises the explainability of the final predictions. Therefore, as with the introduction of any novel technology, it's crucial to implement safety protocols to mitigate misuse. This includes mechanisms for ensuring the accuracy of inputs and establishing procedures to comprehend the criteria the model employs for predictions. By doing so, we can improve the model's reliability and mitigate potential issues.

\section{Qualitative Results}
\label{sec:qualitative}

\subsection{More Visualizations}

We provide more visualizations of panoptic segmentation, part segmentation and referring segmentation in \cref{fig:demo-appendix,fig:demo-appendix-two}. 

\subsection{Combining with SAM}
We integrate our model with the mask outputs generated by the ViT-H Image encoder from Segment Anything (SAM) \cite{kirillov2023segment}. The encoder is trained on SA-1B which encompasses a broad spectrum of objects and masks within each image, enabling us to enhance our segmentation output by utilizing the high-quality masks from the SAM encoder to generate finer, more detailed masks.

To elaborate, in the context of panoptic segmentation, we implement a voting scheme between our pixel-wise annotations and the masks from Segment Anything (SAM), enriching these masks with our labels. For objects where our model demonstrates a strong understanding of hierarchy, such as "person" or "bird", we substitute the SAM masks with ours. This approach enables us to optimize hierarchical outcomes in the face of highly complex images.

\begin{figure*}
  \centering
  \includegraphics[width=1.0\textwidth]{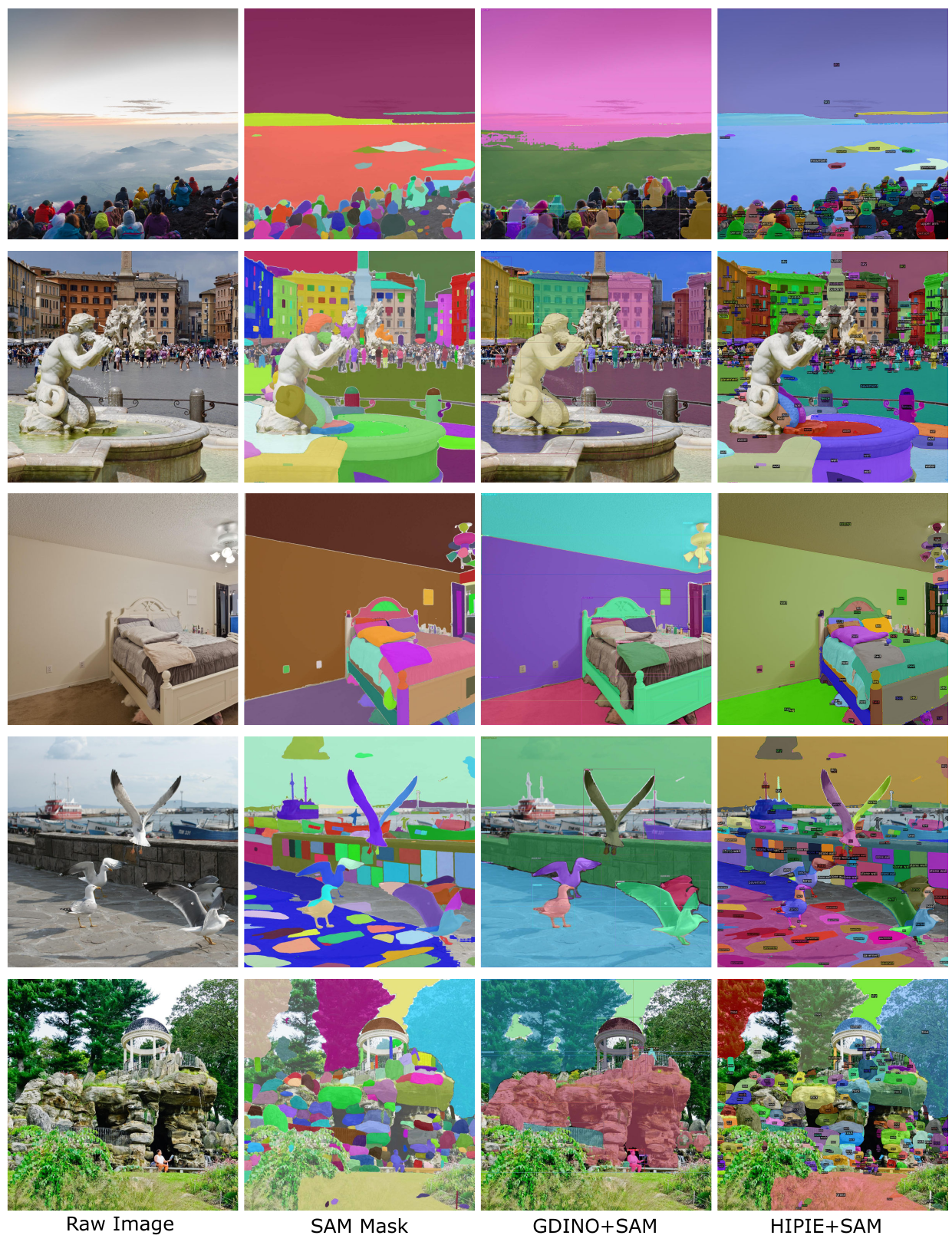}
  \caption{
   Additional results of merging \ours with SAM for hierarchical segmentation. By integrating the part masks from our model and conducting a vote among SAM's panoptic masks, we generate finely detailed mask outputs. Our method demonstrates fewer misclassifications and overlooked masks across the SA-1B dataset compared to the Grounding DINO + SAM approach. Furthermore, our technique excels in differentiating between intra-class objects and identifying distinct object parts.
  }
  \label{fig:demo-sam2}
\end{figure*}

\begin{figure*}
  \centering
  \includegraphics[width=1.0\textwidth]{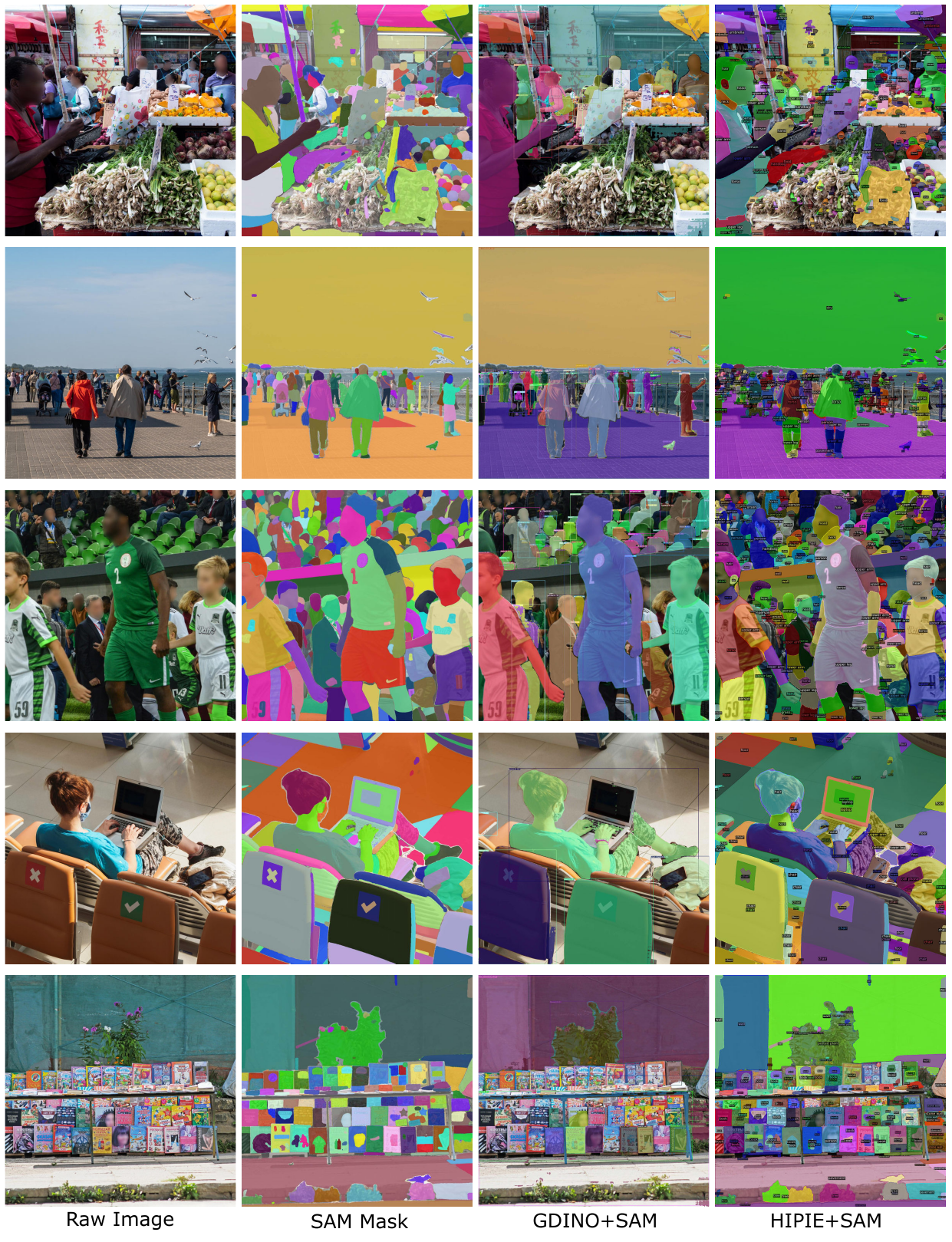}
  \caption{
   Additional results of merging \ours with SAM for hierarchical segmentation. By integrating the part masks from our model and conducting a vote among SAM's panoptic masks, we generate finely detailed mask outputs. Our method demonstrates fewer misclassifications and overlooked masks across the SA-1B dataset compared to the Grounding DINO + SAM approach. Furthermore, our technique excels in differentiating between intra-class objects and identifying distinct object parts.
  }
  \label{fig:demo-sam3}
\end{figure*}

\begin{figure*}
  \centering
  \includegraphics[width=1.0\textwidth]{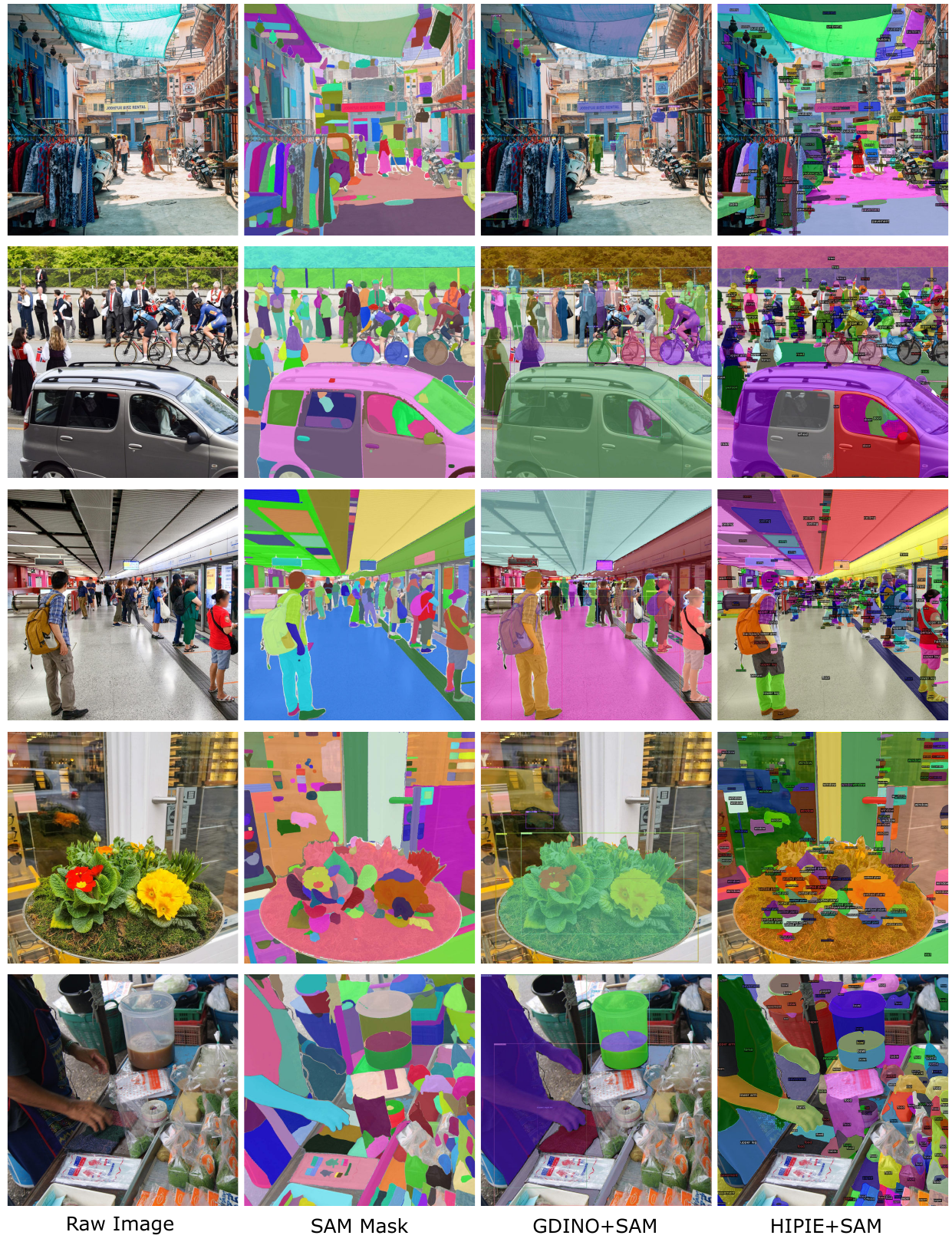}
  \caption{
   Additional results of merging \ours with SAM for hierarchical segmentation. By integrating the part masks from our model and conducting a vote among SAM's panoptic masks, we generate finely detailed mask outputs. Our method demonstrates fewer misclassifications and overlooked masks across the SA-1B dataset compared to the Grounding DINO + SAM approach. Furthermore, our technique excels in differentiating between intra-class objects and identifying distinct object parts.
  }
  \label{fig:demo-sam5}
\end{figure*}

Based on our observations from the figures, it's evident that Grounding DINO generates instance segmentation bounding boxes and subsequently uses SAM for the application of the segmentation masks. While this method proves effective for most datasets, SA-1B is a highly complex set featuring a vast array of whole objects, parts and subparts. Our qualitative findings suggest that the a single granularity instance segmentation model may fail to fully capture all objects/parts within an image or may incorrectly identify them. This consequently leads to SAM receiving sub-optimal bounding boxes for segmentation, resulting in fewer and less accurate masks (see third columns in \cref{fig:demo-sam2,fig:demo-sam3,fig:demo-sam5}). In contrast, our methodology (see last columns in \cref{fig:demo-sam2,fig:demo-sam3,fig:demo-sam5}) integrates the SAM encoder masks with our annotations and hierarchical masks wherever feasible. This results in a significantly more fine-grained and accurate output, proving superior in handling complex datasets such as SA-1B.

\subsection{Combining with Stable Diffusion}
\begin{figure*}
  \centering
  \includegraphics[width=1.0\textwidth]{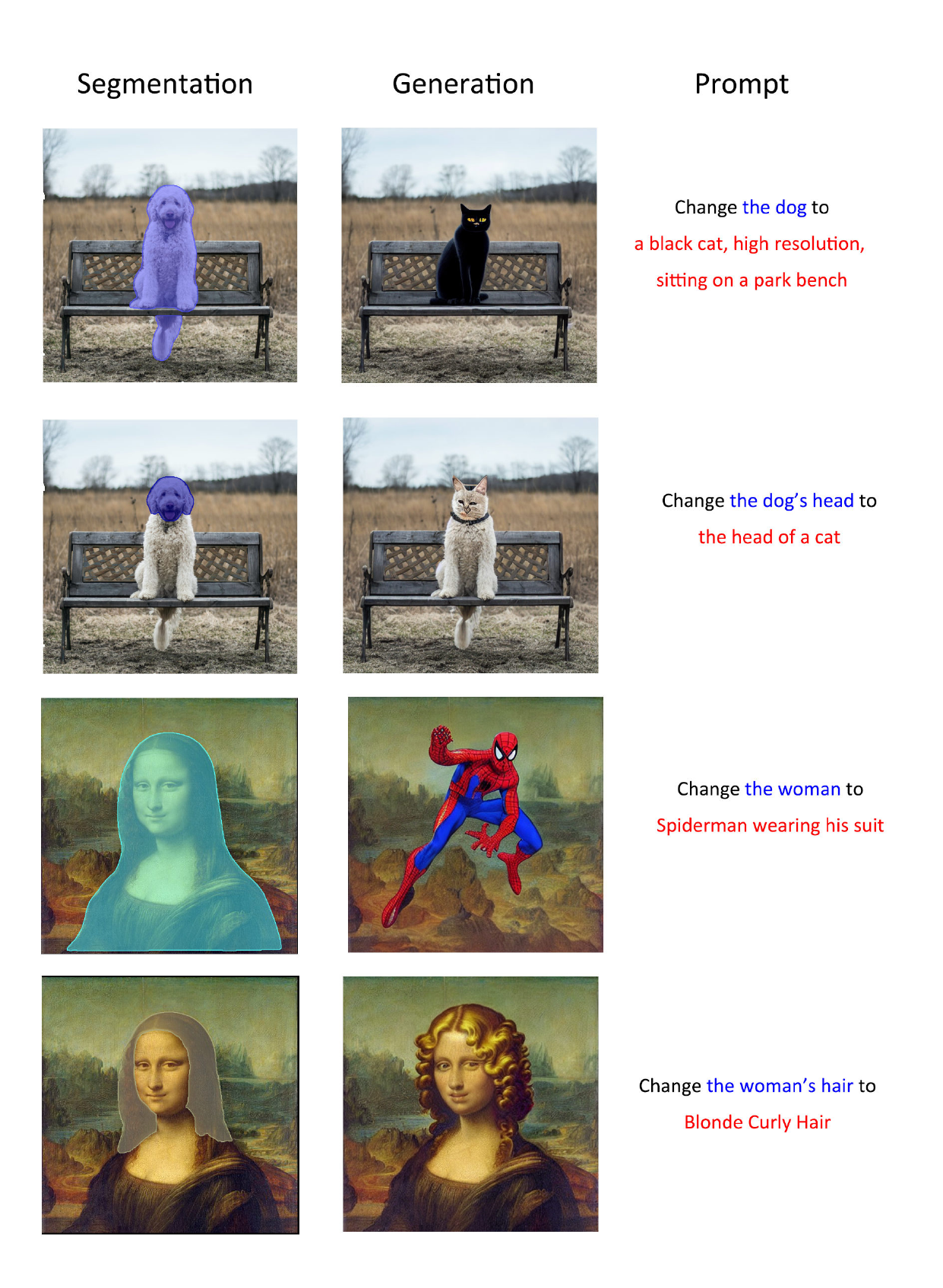}
  \caption{
   Results of combining \ours with Stable Diffusion for Image inpainting. We leverage our segmentation model to generate masks for the redrawing process. Our model can uniquely achieve fine-grained control by providing part segmentation masks. 
  }
  \label{fig:demo-diffusion}
\end{figure*}

As an interesting experiment, we combined our model with image generation model Stable-Diffusion\cite{rombach2021highresolution} in \cref{fig:demo-diffusion}.  Given a source expression and target prompt, we first use \ours's segmentation capability to find the corresponding masks, which are then used for image inpainting. Notably, our model can uniquely achieve fine-grained control over object parts by providing part segmentation masks.

\figDemoAppendix{h}
\figDemoAppendixTwo{h}
\captionsetup[sub]{font=small}

%% file: main.bbl
\begin{thebibliography}{10}

\bibitem{adelson2001seeing}
E.~H. Adelson.
\newblock On seeing stuff: the perception of materials by humans and machines.
\newblock In {\em Human vision and electronic imaging VI}, volume 4299, pages 1--12. SPIE, 2001.

\bibitem{bucher2019zero}
M.~Bucher, T.-H. Vu, M.~Cord, and P.~P{\'e}rez.
\newblock Zero-shot semantic segmentation.
\newblock In {\em NeurIPS}, 2019.

\bibitem{cai2019cascade}
Z.~Cai and N.~Vasconcelos.
\newblock Cascade r-cnn: high quality object detection and instance segmentation.
\newblock {\em IEEE transactions on pattern analysis and machine intelligence}, 43(5):1483--1498, 2019.

\bibitem{cheng2021mask2former}
B.~Cheng, I.~Misra, A.~G. Schwing, A.~Kirillov, and R.~Girdhar.
\newblock Masked-attention mask transformer for universal image segmentation.
\newblock In {\em Proceedings of the IEEE/CVF conference on computer vision and pattern recognition}, pages 1290--1299, 2022.

\bibitem{de2021part}
D.~de~Geus, P.~Meletis, C.~Lu, X.~Wen, and G.~Dubbelman.
\newblock Part-aware panoptic segmentation.
\newblock In {\em Proceedings of the IEEE/CVF Conference on Computer Vision and Pattern Recognition}, pages 5485--5494, 2021.

\bibitem{devlin2018bert}
J.~Devlin, M.-W. Chang, K.~Lee, and K.~Toutanova.
\newblock Bert: Pre-training of deep bidirectional transformers for language understanding.
\newblock {\em arXiv preprint arXiv:1810.04805}, 2018.

\bibitem{dhanachandra2015image}
N.~Dhanachandra, K.~Manglem, and Y.~J. Chanu.
\newblock Image segmentation using k-means clustering algorithm and subtractive clustering algorithm.
\newblock {\em Procedia Computer Science}, 54:764--771, 2015.

\bibitem{ding2021vision}
H.~Ding, C.~Liu, S.~Wang, and X.~Jiang.
\newblock Vision-language transformer and query generation for referring segmentation.
\newblock In {\em Proceedings of the IEEE/CVF International Conference on Computer Vision}, pages 16321--16330, 2021.

\bibitem{ding2022vlt}
H.~Ding, C.~Liu, S.~Wang, and X.~Jiang.
\newblock Vlt: Vision-language transformer and query generation for referring segmentation.
\newblock {\em IEEE Transactions on Pattern Analysis and Machine Intelligence}, 2022.

\bibitem{ding2022open}
Z.~Ding, J.~Wang, and Z.~Tu.
\newblock Open-vocabulary panoptic segmentation with maskclip.
\newblock {\em arXiv preprint arXiv:2208.08984}, 2022.

\bibitem{dosovitskiyimage}
A.~Dosovitskiy, L.~Beyer, A.~Kolesnikov, D.~Weissenborn, X.~Zhai, T.~Unterthiner, M.~Dehghani, M.~Minderer, G.~Heigold, S.~Gelly, et~al.
\newblock An image is worth 16x16 words: Transformers for image recognition at scale.
\newblock In {\em International Conference on Learning Representations}, 2020.

\bibitem{everingham2010pascal}
M.~Everingham, L.~Van~Gool, C.~K. Williams, J.~Winn, and A.~Zisserman.
\newblock The pascal visual object classes (voc) challenge.
\newblock {\em International journal of computer vision}, 88:303--338, 2010.

\bibitem{feng2021encoder}
G.~Feng, Z.~Hu, L.~Zhang, and H.~Lu.
\newblock Encoder fusion network with co-attention embedding for referring image segmentation.
\newblock In {\em Proceedings of the IEEE/CVF Conference on Computer Vision and Pattern Recognition}, pages 15506--15515, 2021.

\bibitem{forsyth2002computer}
D.~A. Forsyth and J.~Ponce.
\newblock {\em Computer vision: a modern approach}.
\newblock prentice hall professional technical reference, 2002.

\bibitem{ge2021yolox}
Z.~Ge, S.~Liu, F.~Wang, Z.~Li, and J.~Sun.
\newblock Yolox: Exceeding yolo series in 2021.
\newblock {\em arXiv preprint arXiv:2107.08430}, 2021.

\bibitem{ghiasi2022scaling}
G.~Ghiasi, X.~Gu, Y.~Cui, and T.-Y. Lin.
\newblock Scaling open-vocabulary image segmentation with image-level labels.
\newblock In {\em Computer Vision--ECCV 2022: 17th European Conference, Tel Aviv, Israel, October 23--27, 2022, Proceedings, Part XXXVI}, pages 540--557. Springer, 2022.

\bibitem{harary2022unsupervised}
S.~Harary, E.~Schwartz, A.~Arbelle, P.~Staar, S.~Abu-Hussein, E.~Amrani, R.~Herzig, A.~Alfassy, R.~Giryes, H.~Kuehne, et~al.
\newblock Unsupervised domain generalization by learning a bridge across domains.
\newblock In {\em Proceedings of the IEEE/CVF Conference on Computer Vision and Pattern Recognition}, pages 5280--5290, 2022.

\bibitem{he2022masked}
K.~He, X.~Chen, S.~Xie, Y.~Li, P.~Doll{\'a}r, and R.~Girshick.
\newblock Masked autoencoders are scalable vision learners.
\newblock In {\em Proceedings of the IEEE/CVF Conference on Computer Vision and Pattern Recognition}, pages 16000--16009, 2022.

\bibitem{he2016deep}
K.~He, X.~Zhang, S.~Ren, and J.~Sun.
\newblock Deep residual learning for image recognition.
\newblock In {\em Proceedings of the IEEE conference on computer vision and pattern recognition}, pages 770--778, 2016.

\bibitem{hu2016segmentation}
R.~Hu, M.~Rohrbach, and T.~Darrell.
\newblock Segmentation from natural language expressions.
\newblock In {\em Proceedings of the European Conference on Computer Vision (ECCV)}, 2016.

\bibitem{hui2020linguistic}
T.~Hui, S.~Liu, S.~Huang, G.~Li, S.~Yu, F.~Zhang, and J.~Han.
\newblock Linguistic structure guided context modeling for referring image segmentation.
\newblock In {\em Computer Vision--ECCV 2020: 16th European Conference, Glasgow, UK, August 23--28, 2020, Proceedings, Part X 16}, pages 59--75. Springer, 2020.

\bibitem{jagadeesh2022multijppf}
S.~K. Jagadeesh, R.~Schuster, and D.~Stricker.
\newblock Multi-task fusion for efficient panoptic-part segmentation.
\newblock {\em arXiv preprint arXiv:2212.07671}, 2022.

\bibitem{jing2021locate}
Y.~Jing, T.~Kong, W.~Wang, L.~Wang, L.~Li, and T.~Tan.
\newblock Locate then segment: A strong pipeline for referring image segmentation.
\newblock In {\em Proceedings of the IEEE/CVF Conference on Computer Vision and Pattern Recognition}, pages 9858--9867, 2021.

\bibitem{kirillov2019panoptic}
A.~Kirillov, K.~He, R.~Girshick, C.~Rother, and P.~Doll{\'a}r.
\newblock Panoptic segmentation.
\newblock In {\em Proceedings of the IEEE/CVF Conference on Computer Vision and Pattern Recognition}, pages 9404--9413, 2019.

\bibitem{kirillov2023segment}
A.~Kirillov, E.~Mintun, N.~Ravi, H.~Mao, C.~Rolland, L.~Gustafson, T.~Xiao, S.~Whitehead, A.~C. Berg, W.-Y. Lo, et~al.
\newblock Segment anything.
\newblock {\em arXiv preprint arXiv:2304.02643}, 2023.

\bibitem{kuhn1955hungarian}
H.~W. Kuhn.
\newblock The hungarian method for the assignment problem.
\newblock {\em Naval research logistics quarterly}, 2(1-2):83--97, 1955.

\bibitem{li2022languagedriven}
B.~Li, K.~Q. Weinberger, S.~Belongie, V.~Koltun, and R.~Ranftl.
\newblock Language-driven semantic segmentation.
\newblock In {\em International Conference on Learning Representations}, 2022.

\bibitem{li2022dn}
F.~Li, H.~Zhang, S.~Liu, J.~Guo, L.~M. Ni, and L.~Zhang.
\newblock Dn-detr: Accelerate detr training by introducing query denoising.
\newblock In {\em Proceedings of the IEEE/CVF Conference on Computer Vision and Pattern Recognition}, pages 13619--13627, 2022.

\bibitem{li2022mask}
F.~Li, H.~Zhang, H.~xu, S.~Liu, L.~Zhang, L.~M. Ni, and H.-Y. Shum.
\newblock Mask dino: Towards a unified transformer-based framework for object detection and segmentation, 2022.

\bibitem{li2021grounded}
L.~H. Li*, P.~Zhang*, H.~Zhang*, J.~Yang, C.~Li, Y.~Zhong, L.~Wang, L.~Yuan, L.~Zhang, J.-N. Hwang, K.-W. Chang, and J.~Gao.
\newblock Grounded language-image pre-training.
\newblock In {\em CVPR}, 2022.

\bibitem{li2021referring}
M.~Li and L.~Sigal.
\newblock Referring transformer: A one-step approach to multi-task visual grounding.
\newblock {\em Advances in neural information processing systems}, 34:19652--19664, 2021.

\bibitem{li2022exploring}
Y.~Li, H.~Mao, R.~Girshick, and K.~He.
\newblock Exploring plain vision transformer backbones for object detection.
\newblock In {\em Computer Vision--ECCV 2022: 17th European Conference, Tel Aviv, Israel, October 23--27, 2022, Proceedings, Part IX}, pages 280--296. Springer, 2022.

\bibitem{liang2022open}
F.~Liang, B.~Wu, X.~Dai, K.~Li, Y.~Zhao, H.~Zhang, P.~Zhang, P.~Vajda, and D.~Marculescu.
\newblock Open-vocabulary semantic segmentation with mask-adapted clip.
\newblock {\em arXiv preprint arXiv:2210.04150}, 2022.

\bibitem{lin2017focal}
T.-Y. Lin, P.~Goyal, R.~Girshick, K.~He, and P.~Doll{\'a}r.
\newblock Focal loss for dense object detection.
\newblock In {\em Proceedings of the IEEE international conference on computer vision}, pages 2980--2988, 2017.

\bibitem{lin2014microsoft}
T.-Y. Lin, M.~Maire, S.~Belongie, J.~Hays, P.~Perona, D.~Ramanan, P.~Doll{\'a}r, and C.~L. Zitnick.
\newblock Microsoft coco: Common objects in context.
\newblock In {\em Computer Vision--ECCV 2014: 13th European Conference, Zurich, Switzerland, September 6-12, 2014, Proceedings, Part V 13}, pages 740--755. Springer, 2014.

\bibitem{liu2023polyformer}
J.~Liu, H.~Ding, Z.~Cai, Y.~Zhang, R.~K. Satzoda, V.~Mahadevan, and R.~Manmatha.
\newblock Polyformer: Referring image segmentation as sequential polygon generation.
\newblock {\em arXiv preprint arXiv:2302.07387}, 2023.

\bibitem{liu2023grounding}
S.~Liu, Z.~Zeng, T.~Ren, F.~Li, H.~Zhang, J.~Yang, C.~Li, J.~Yang, H.~Su, J.~Zhu, et~al.
\newblock Grounding dino: Marrying dino with grounded pre-training for open-set object detection.
\newblock {\em arXiv preprint arXiv:2303.05499}, 2023.

\bibitem{long2015fully}
J.~Long, E.~Shelhamer, and T.~Darrell.
\newblock Fully convolutional networks for semantic segmentation.
\newblock In {\em Proceedings of the IEEE conference on computer vision and pattern recognition}, pages 3431--3440, 2015.

\bibitem{loshchilovdecoupled}
I.~Loshchilov and F.~Hutter.
\newblock Decoupled weight decay regularization.
\newblock In {\em International Conference on Learning Representations}, 2017.

\bibitem{mottaghi2014role}
R.~Mottaghi, X.~Chen, X.~Liu, N.-G. Cho, S.-W. Lee, S.~Fidler, R.~Urtasun, and A.~Yuille.
\newblock The role of context for object detection and semantic segmentation in the wild.
\newblock In {\em Proceedings of the IEEE conference on computer vision and pattern recognition}, pages 891--898, 2014.

\bibitem{muchen2021referring}
L.~Muchen and S.~Leonid.
\newblock Referring transformer: A one-step approach to multi-task visual grounding.
\newblock In {\em Thirty-Fifth Conference on Neural Information Processing Systems}, 2021.

\bibitem{nagaraja2016modeling}
V.~K. Nagaraja, V.~I. Morariu, and L.~S. Davis.
\newblock Modeling context between objects for referring expression understanding.
\newblock In {\em Computer Vision--ECCV 2016: 14th European Conference, Amsterdam, The Netherlands, October 11--14, 2016, Proceedings, Part IV 14}, pages 792--807. Springer, 2016.

\bibitem{nock2004statistical}
R.~Nock and F.~Nielsen.
\newblock Statistical region merging.
\newblock {\em IEEE Transactions on pattern analysis and machine intelligence}, 26(11):1452--1458, 2004.

\bibitem{radford2021learning}
A.~Radford, J.~W. Kim, C.~Hallacy, A.~Ramesh, G.~Goh, S.~Agarwal, G.~Sastry, A.~Askell, P.~Mishkin, J.~Clark, et~al.
\newblock Learning transferable visual models from natural language supervision.
\newblock In {\em International conference on machine learning}, pages 8748--8763. PMLR, 2021.

\bibitem{rao2022denseclip}
Y.~Rao, W.~Zhao, G.~Chen, Y.~Tang, Z.~Zhu, G.~Huang, J.~Zhou, and J.~Lu.
\newblock Denseclip: Language-guided dense prediction with context-aware prompting.
\newblock In {\em Proceedings of the IEEE/CVF Conference on Computer Vision and Pattern Recognition}, pages 18082--18091, 2022.

\bibitem{rezatofighi2019generalized}
H.~Rezatofighi, N.~Tsoi, J.~Gwak, A.~Sadeghian, I.~Reid, and S.~Savarese.
\newblock Generalized intersection over union: A metric and a loss for bounding box regression.
\newblock In {\em Proceedings of the IEEE/CVF conference on computer vision and pattern recognition}, pages 658--666, 2019.

\bibitem{rombach2021highresolution}
R.~Rombach, A.~Blattmann, D.~Lorenz, P.~Esser, and B.~Ommer.
\newblock High-resolution image synthesis with latent diffusion models, 2021.

\bibitem{rombach2022high}
R.~Rombach, A.~Blattmann, D.~Lorenz, P.~Esser, and B.~Ommer.
\newblock High-resolution image synthesis with latent diffusion models.
\newblock In {\em Proceedings of the IEEE/CVF Conference on Computer Vision and Pattern Recognition}, pages 10684--10695, 2022.

\bibitem{shao2019objects365}
S.~Shao, Z.~Li, T.~Zhang, C.~Peng, G.~Yu, X.~Zhang, J.~Li, and J.~Sun.
\newblock Objects365: A large-scale, high-quality dataset for object detection.
\newblock In {\em Proceedings of the IEEE/CVF international conference on computer vision}, pages 8430--8439, 2019.

\bibitem{sudre2017generalised}
C.~H. Sudre, W.~Li, T.~Vercauteren, S.~Ourselin, and M.~Jorge~Cardoso.
\newblock Generalised dice overlap as a deep learning loss function for highly unbalanced segmentations.
\newblock In {\em Deep Learning in Medical Image Analysis and Multimodal Learning for Clinical Decision Support: Third International Workshop, DLMIA 2017, and 7th International Workshop, ML-CDS 2017, Held in Conjunction with MICCAI 2017, Qu{\'e}bec City, QC, Canada, September 14, Proceedings 3}, pages 240--248. Springer, 2017.

\bibitem{szeliski2022computer}
R.~Szeliski.
\newblock {\em Computer vision: algorithms and applications}.
\newblock Springer Nature, 2022.

\bibitem{wang2022cris}
Z.~Wang, Y.~Lu, Q.~Li, X.~Tao, Y.~Guo, M.~Gong, and T.~Liu.
\newblock Cris: Clip-driven referring image segmentation.
\newblock In {\em Proceedings of the IEEE/CVF conference on computer vision and pattern recognition}, pages 11686--11695, 2022.

\bibitem{wu2022towards}
J.~Wu, X.~Li, X.~Li, H.~Ding, Y.~Tong, and D.~Tao.
\newblock Towards robust referring image segmentation.
\newblock {\em arXiv preprint arXiv:2209.09554}, 2022.

\bibitem{xian2019semantic}
Y.~Xian, S.~Choudhury, Y.~He, B.~Schiele, and Z.~Akata.
\newblock Semantic projection network for zero-and few-label semantic segmentation.
\newblock In {\em Proceedings of the IEEE/CVF Conference on Computer Vision and Pattern Recognition}, pages 8256--8265, 2019.

\bibitem{xu2022groupvit}
J.~Xu, S.~De~Mello, S.~Liu, W.~Byeon, T.~Breuel, J.~Kautz, and X.~Wang.
\newblock Groupvit: Semantic segmentation emerges from text supervision.
\newblock In {\em Proceedings of the IEEE/CVF Conference on Computer Vision and Pattern Recognition}, pages 18134--18144, 2022.

\bibitem{xu2023learning}
J.~Xu, J.~Hou, Y.~Zhang, R.~Feng, Y.~Wang, Y.~Qiao, and W.~Xie.
\newblock Learning open-vocabulary semantic segmentation models from natural language supervision.
\newblock {\em arXiv preprint arXiv:2301.09121}, 2023.

\bibitem{xu2023open}
J.~Xu, S.~Liu, A.~Vahdat, W.~Byeon, X.~Wang, and S.~De~Mello.
\newblock Open-vocabulary panoptic segmentation with text-to-image diffusion models.
\newblock {\em arXiv preprint arXiv:2303.04803}, 2023.

\bibitem{xu2022simple}
M.~Xu, Z.~Zhang, F.~Wei, Y.~Lin, Y.~Cao, H.~Hu, and X.~Bai.
\newblock A simple baseline for open-vocabulary semantic segmentation with pre-trained vision-language model.
\newblock In {\em Computer Vision--ECCV 2022: 17th European Conference, Tel Aviv, Israel, October 23--27, 2022, Proceedings, Part XXIX}, pages 736--753. Springer, 2022.

\bibitem{yan2023universal}
B.~Yan, Y.~Jiang, J.~Wu, D.~Wang, P.~Luo, Z.~Yuan, and H.~Lu.
\newblock Universal instance perception as object discovery and retrieval.
\newblock {\em arXiv preprint arXiv:2303.06674}, 2023.

\bibitem{yang2022lavt}
Z.~Yang, J.~Wang, Y.~Tang, K.~Chen, H.~Zhao, and P.~H. Torr.
\newblock Lavt: Language-aware vision transformer for referring image segmentation.
\newblock In {\em Proceedings of the IEEE/CVF Conference on Computer Vision and Pattern Recognition}, pages 18155--18165, 2022.

\bibitem{yu2018mattnet}
L.~Yu, Z.~Lin, X.~Shen, J.~Yang, X.~Lu, M.~Bansal, and T.~L. Berg.
\newblock Mattnet: Modular attention network for referring expression comprehension.
\newblock In {\em CVPR}, 2018.

\bibitem{yu2016modeling}
L.~Yu, P.~Poirson, S.~Yang, A.~C. Berg, and T.~L. Berg.
\newblock Modeling context in referring expressions.
\newblock In {\em Computer Vision--ECCV 2016: 14th European Conference, Amsterdam, The Netherlands, October 11-14, 2016, Proceedings, Part II 14}, pages 69--85. Springer, 2016.

\bibitem{zhang2022dino}
H.~Zhang, F.~Li, S.~Liu, L.~Zhang, H.~Su, J.~Zhu, L.~M. Ni, and H.-Y. Shum.
\newblock Dino: Detr with improved denoising anchor boxes for end-to-end object detection.
\newblock {\em arXiv preprint arXiv:2203.03605}, 2022.

\bibitem{zhao2023unleashing}
W.~Zhao, Y.~Rao, Z.~Liu, B.~Liu, J.~Zhou, and J.~Lu.
\newblock Unleashing text-to-image diffusion models for visual perception.
\newblock {\em arXiv preprint arXiv:2303.02153}, 2023.

\bibitem{zhou2019semantic}
B.~Zhou, H.~Zhao, X.~Puig, T.~Xiao, S.~Fidler, A.~Barriuso, and A.~Torralba.
\newblock Semantic understanding of scenes through the ade20k dataset.
\newblock {\em International Journal of Computer Vision}, 127:302--321, 2019.

\bibitem{zhu2020deformable}
X.~Zhu, W.~Su, L.~Lu, B.~Li, X.~Wang, and J.~Dai.
\newblock Deformable detr: Deformable transformers for end-to-end object detection.
\newblock {\em arXiv preprint arXiv:2010.04159}, 2020.

\bibitem{zou2022generalized}
X.~Zou, Z.-Y. Dou, J.~Yang, Z.~Gan, L.~Li, C.~Li, X.~Dai, H.~Behl, J.~Wang, L.~Yuan, et~al.
\newblock Generalized decoding for pixel, image, and language.
\newblock {\em arXiv preprint arXiv:2212.11270}, 2022.

\bibitem{zou2023segment}
X.~Zou, J.~Yang, H.~Zhang, F.~Li, L.~Li, J.~Gao, and Y.~J. Lee.
\newblock Segment everything everywhere all at once.
\newblock {\em arXiv preprint arXiv:2304.06718}, 2023.

\end{thebibliography}
